\documentclass[lettersize,journal]{IEEEtran}
\usepackage{epsfig,rotating,setspace,latexsym,amsmath,epsf,amssymb,amsfonts,bm,theorem,cite,algorithm,algorithmic,graphicx,epsf,authblk,epstopdf,url,color,mwe,xcolor,multirow, booktabs,threeparttable}
\usepackage{amsmath,amsfonts}
\usepackage{algorithmic}
\usepackage{algorithm}
\usepackage{array}
\usepackage[caption=false,font=normalsize,labelfont=sf,textfont=sf]{subfig}
\usepackage{textcomp}
\usepackage{stfloats}
\usepackage{url}
\usepackage{verbatim}
\usepackage{graphicx}
\usepackage{cite}
\begin{document}

\title{Number Systems for Deep Neural Network Architectures: A Survey}

\author[1]{Ghada Alsuhli}
\author[1]{Vasileios Sakellariou}

\author[1]{Hani Saleh,~\IEEEmembership{Senior Member,~IEEE,}}
\author[1]{Mahmoud Al-Qutayri,~\IEEEmembership{Senior Member,~IEEE,}}

\author[1]{Baker Mohammad,~\IEEEmembership{Senior Member,~IEEE,}}
\author[1]{Thanos Stouraitis}

\affil[1]{Department of Electrical Engineering and Computer Science, System on Chip Center, Khalifa University}




\maketitle

\begin{abstract}
Deep neural networks (DNNs) have become an enabling component for a myriad of artificial intelligence applications. DNNs have shown sometimes superior performance, even compared to humans, in cases such as self-driving, health applications, etc. Because of their computational complexity, deploying DNNs in resource-constrained devices still faces many challenges related to computing complexity, energy efficiency, latency, and cost. To this end, several research directions are being pursued by both academia and industry to accelerate and efficiently implement DNNs. One important direction is determining the appropriate data representation for the massive amount of data involved in DNN processing. Using conventional number systems has been found to be sub-optimal for DNNs. Alternatively, a great body of research focuses on exploring suitable number systems. This article aims to provide a comprehensive survey and discussion about alternative number systems for more efficient representations of DNN data. Various number systems (conventional/unconventional) exploited for DNNs are discussed. The impact of these number systems on the performance and hardware design of DNNs is considered. In addition, this paper highlights the challenges associated with each number system and various solutions that are proposed for addressing them. The reader will be able to understand the importance of an efficient number system for DNN, learn about the widely used number systems for DNN, understand the trade-offs between various number systems, and consider various design aspects that affect the impact of number systems on DNN performance. In addition, the recent trends and related research opportunities will be highlighted.
\end{abstract}


\section{Introduction}
\IEEEPARstart{D}{uring} the past decade, Deep Neural Networks (DNNs) have shown outstanding performance in a myriad of Artificial Intelligence (AI) applications. Since their success in speech recognition \cite {deng2013recent} and image recognition \cite{krizhevsky2017imagenet}, great attention has been drawn to DNNs from academia and industry \cite{guo2018survey}. Although DNNs are inspired by the deep hierarchical structures of the human brain, they have exceeded the human accuracy in a number of domains \cite{sze2017efficient}. Nowadays, the contribution of DNNs is notable in many fields including self-driving cars \cite{gupta2021deep}, speech recognition \cite{shewalkar2019performance}, computer vision \cite{buhrmester2021analysis}, natural language processing \cite{otter2020survey}, and medical applications \cite{pustokhina2020effective}. This DNN revolution is helped by the massive accumulation of data and the rapid growth in computing power \cite{ALAM2020302}. 

Because of their high computational complexity and memory space requirements, general-purpose compute engines (like powerful central processing units (CPUs) and Graphics Processing Units (GPUs)), or customized hardware (e.g., using FPGAs or ASICs) have been used to accelerate DNN processing \cite{lecun20191}. While general-purpose compute engines remain dominant for processing DNNs within academia, the industrial applications of DNNs often require implementation on resource-constrained edge devices ( e.g., smartphones or wearable devices) \cite{guo2018survey}. Whether DNNs are run on GPUs or dedicated accelerators, speeding up and/or increasing DNN hardware efficiency without sacrificing their accuracy continues to be a demanding task. The literature includes a large number of survey papers that have been dedicated to highlighting the directions that can be followed to reach these goals \cite{sze2017efficient, gholami2021survey, guo2018survey, wu2021accelerating,ghimire2022survey}. Some examples of these directions are DNN model compression \cite{choudhary2020comprehensive}, quantization \cite{gholami2021survey}, and DNN efficient processing \cite{sze2017efficient}. One of the directions that have a great impact on the performance of DNNs, but has not been comprehensively surveyed yet is the DNN number representation.

As the compute engines use a limited number of bits to represent values, real numbers cannot be infinitely represented. The mapping between a real number and the bits that represent it is called number representation \cite{gohil2021fixed}. Generally speaking, number representation has a great impact on the performance of both general-purpose and customized compute engines. Recalling the huge amount of data that need to be processed in the context of DNNs, the choice of the format used to represent these data is a key factor in determining the precision of DNN data, storage requirements, memory communication, and arithmetic hardware implementation \cite{darvish2020pushing}. This in turn shapes different metrics of the DNN architecture performance; mainly the accuracy, power consumption, throughput, latency, and cost \cite{sze2020efficient}. 

To this end, there is a significant body of literature that has focused on assessing the suitability of specific number systems for DNNs, modifying conventional number systems to fit DNN workloads, or proposing new number systems tailored for DNNs. Some of the leading companies, such as Google \cite{wang2019bfloat16}, NVIDIA \cite{choquette2021nvidia}, Microsoft \cite{darvish2020pushing}, IBM \cite{gupta2015deep}, and Intel \cite{kalamkar2019study,koster2017flexpoint, popescu2018flexpoint}, have contributed in advancing the research in this field. A comprehensive survey of these works will be helpful to furthering the research in this field.  

\begin{figure*}[!t]
\centering
\subfloat[]{\includegraphics[trim={0cm 4cm 0 0},width=0.43 \linewidth]{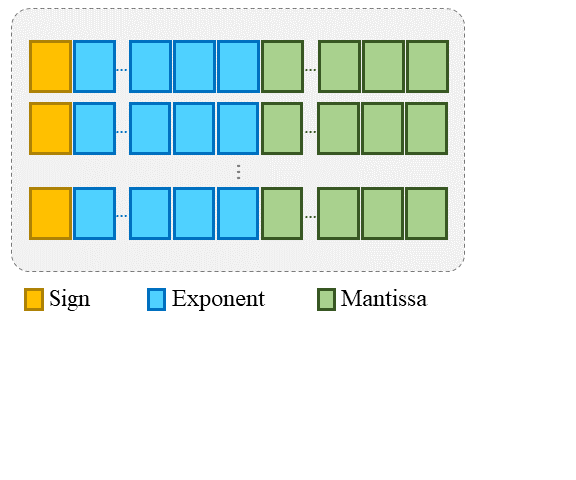}%
\label{fp_rep}}
\hfil
\subfloat[]{\includegraphics[trim={0cm 4cm 0 0},width=0.43 \linewidth]{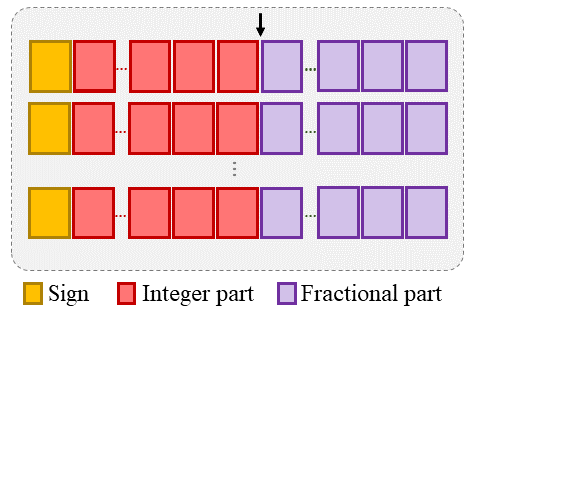}%
\label{fxp_rep}}
\hfil
\subfloat[]{\includegraphics[trim={0.5cm 0 0 0},width=0.43 \linewidth]{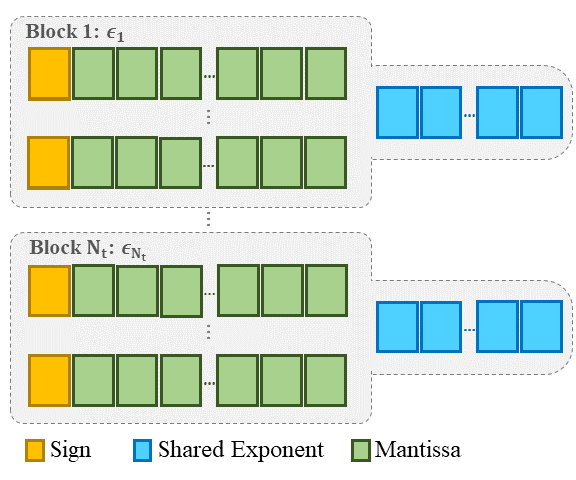}%
\label{bfp_rep}}
\hfil
\subfloat[]{\includegraphics[trim={0.5cm 0 0 0},width=0.43 \linewidth]{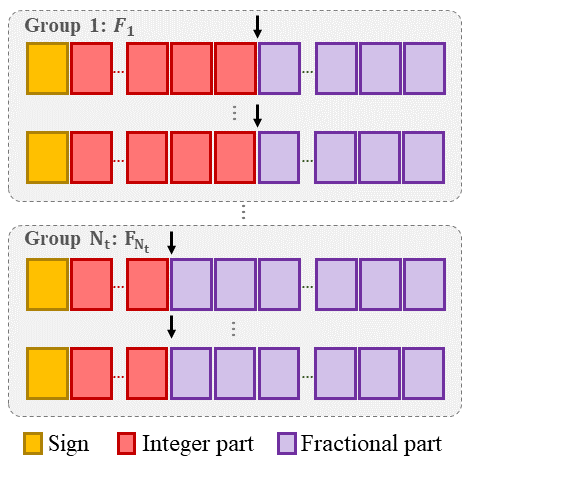}%
\label{dfxp_rep}}
\hfil
\subfloat[]{\includegraphics[trim={0.5cm 4cm 0 0},width=0.43 \linewidth]{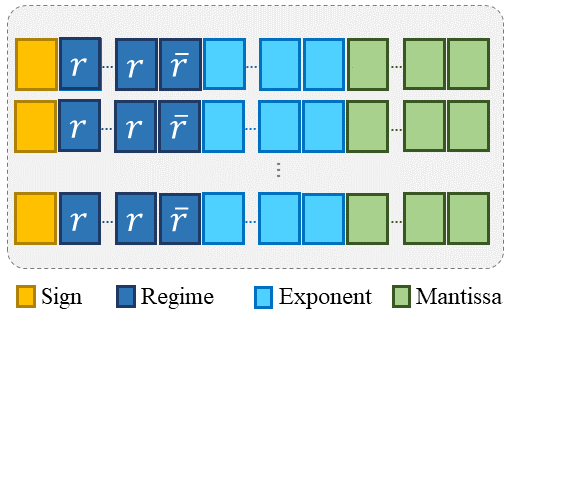}%
\label{Posit_rep}}
\hfil
\subfloat[]{\includegraphics[trim={0.5cm 4cm 0 0},width=0.43 \linewidth]{fxp_rep}%
\label{lns_rep}}
\hfil
\subfloat[]{\includegraphics[width=0.43\linewidth]{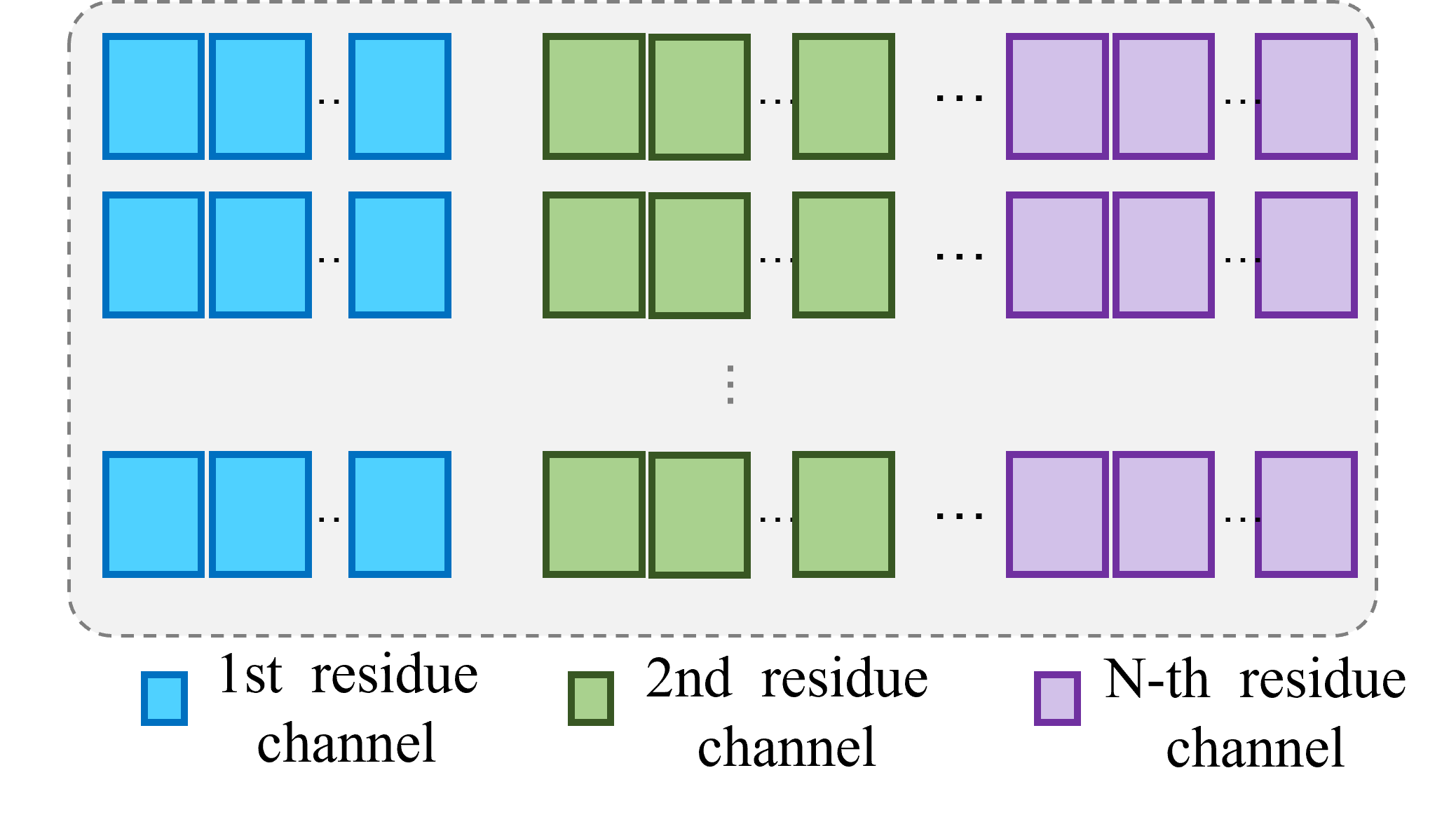}%
\label{fig_first_case}}

\caption{Conventional [(a) FP, (b) FXP]  and unconventional [(c) BFP, (d) DFXP, (e) Posit, (f) LNS, (g) RNS] number systems for DNNs}
\label{all_rep}
\end{figure*}

While conventional number systems like Floating Point (FLP) and Fixed Point (FXP) representations are frequently used for DNN engines, several unconventional number systems are found to be more efficient for DNN implementation. Such alternative number systems are presented in this survey and they are the Logarithmic Number System (LNS), Residue Number System (RNS), Block Floating Point Number System (BFP), Dynamic Fixed Point Number System (DFXP), and Posit Number System (PNS). Figure \ref{all_rep} shows the bit visualization of conventional and unconventional number systems used in DNN implementation. The structure of the survey is summarized as follows. 
\begin{itemize}
    \item Section \ref{CNS} gives an overview of conventional number systems and their utilization for DNNs. 
    \item Section \ref{LNS_Sec} classifies the DNNs that adopt the logarithmic number system.  
    \item Section \ref{RNS_Sec} describes the concepts behind the residue number system and its employment for DNNs.
    \item Section \ref{BFP} describes the block floating point representation and the efforts done to make it suitable for DNNs implementation. 
    \item Section \ref{DFXP_sec} discusses the dynamic fixed point format and the work done to calibrate the parameters associated with this format.  
    \item Section \ref{Posit_sec} explains various DNN architectures that utilize Posits and the advantages and disadvantages associated with these architectures.   
    \item Section \ref{Future} provides an insight into recent trends and research opportunities in the field of DNN number systems. 
\end{itemize}

\section{Conventional Number Systems for DNN architectures}\label{CNS}
The two conventional number systems, mainly the floating point and the fixed point, are the common choice for almost all general-purpose DNN engines. While the FLP representation is usually used for modern computation platforms (e.g., CPUs and GPUs), where high precision is a must, FXP is more common in low-cost computation platforms that are used in applications that demand high speed, low power consumption, and small chip area. In this section, these two representations are introduced and their utilization for implementing DNN hardware is briefly discussed, in order to facilitate the comparison between conventional and unconventional number systems.

\subsection{FLP for DNN Architectures}
In the FLP number system, a number $n$ is represented using a sign (1 bit), an exponent $e$ (unsigned integer of length $es$) and a mantissa $m$ (unsigned integer of length $ms$) (Figure \ref{fp_rep}) and its value is given by
 \begin{equation} \label{fp_format}
{n} =(-1)^s \times  2^{e-e_{max}} \times (1+\frac{m}{2^{ms}}),
\end{equation}
where $e_{max}=2^{es-1}-1$ is a bias used to ease the representation of both negative and Positive exponents. 
Although there are several FLP formats \cite{harris2022hardware}, the IEEE 754 FLP format \cite{ieee2019ieee} is the most common representation used by modern computing platforms \cite{sze2020efficient,courbariaux2014training}. According to IEEE 754, the FLP can be of single, double, or quad-precision depending on the used bit-widths (e.g., for the single-precision FLP the bit-width is 32 bits and $es=8$). The single-precision FLP, also called FLP32, is commonly used as a baseline to evaluate the efficiency of other number representations. Unless otherwise stated, the performance degradation or enhancement is presented in comparison to the FLP32 format in this survey as well.

Multiplication of two FLP numbers is implemented in hardware by adding their exponents, multiplying the mantissas, normalizing the resultant mantissa, and adjusting the exponent of the product \cite{leon2021improving}. FLP addition involves comparing the operand exponents, shifting their mantissas (if the exponents are different), adding the mantissas, normalizing the sum mantissa, and adjusting the sum exponent \cite{harris2022hardware}. Usually, the increased complexity of the FLP32 arithmetic requires using a separate unit called Floating Point Unit (FPU) to perform the FLP calculations \cite{abdelaziz2021rethinking}. The high power consumption and cost of this unit limits its usage within embedded processing units such as FPGAs \cite{hassan2020design}. Consequently, the standard FLP32 is rarely used for building efficient DNN architectures \cite{leon2021improving}. To increase the efficiency of the FLP in DNN architectures several custom FLP formats \cite{wang2019bfloat16, choquette2021nvidia, wu2021low, kang2018short} have been proposed. Also new designs of the FLP arithmetic hardware (mainly the multiplier) have been investigated \cite{lee2020design, leon2021improving}.

The 32-bit FLP representation has a wide dynamic range, beyond what is usually required for DNNs \cite{leon2021improving}, resulting in a low information-per-bit metric, which means an unnecessary increase in power consumption, area, and delay. For this reason, the proposed custom FLP representations mainly have reduced bit-width and a different allocation of the bits to mantissa and exponent, than IEEE 754. The bit-width is reduced to 19 bits in Nvidia’s TensorFloat32 \cite{choquette2021nvidia} and 16 bits in Google’s Brain FLP (bfloat16) \cite{wang2019bfloat16} formats used in DNN training engines. 8-bit FLP has been proposed to target the DNN inference in \cite{wu2021low, kang2018short}. These reduced FLP formats proved their efficiency in replacing FLP32 with comparable accuracy, higher throughput, and smaller hardware footprint. It is worth noting that most of these custom FLP formats are used to represent data stored in memory (i.e., weights and activations), whereas, for internal calculations (e.g.,  accumulation and weight updates), FLP32 is used instead to avoid accuracy degradation \cite{wang2019bfloat16,kang2018short,narang2018mixed}. 

In summary, the standard FLP representation has a massive dynamic range, which makes it a good choice for computationally intensive algorithms that include a wide range of values and require high precision. At the same time, the complex and power-hungry FLP calculations make FLP less attractive for DNN accelerators. This leads to using narrower custom FLP formats which require less hardware and memory footprint while preserving the performance of the standard FLP32. However, the utilization of the FLP format for DNN accelerators is relatively limited and it loses ground to fixed point and other alternative representations.

\subsection{FXP for DNN Architectures}
The power inefficiency of the FLP arithmetic is the main motivation to replace it with the FXP format for designing energy-constrained DNN accelerators. A real number $n$ is represented in FXP with the sign, the integer, and the fraction parts. The fixed point format is usually indicated by $<I,F>$ where $I$ and $F$ correspond to the number of bits allocated to the integer and the fractional parts, respectively. In this paper, we use the notations FXP8, for example, to denote the FXP representation with bit-width equal to 8, i.e., $I+F+1=8$. 

In FXP format, the separation between the integer and the fractional parts is implicit and usually done by specifying a scaling factor that is common for all data. Thus, the FXP number can be treated as an integer and, hence, integer arithmetic is used. Integer arithmetic requires substantially fewer logic gates to be implemented and consumes much less chip area and power, compared to FLP arithmetic. This makes FXP attractive to be used for DNN accelerators on the edge. Moreover, the FXP allows for more reduction in the number of bits resulting in a significant reduction in the power consumption, storage requirements, and memory bandwidth \cite{sze2017efficient}.
 
On the other hand, the dynamic range\footnote{The dynamic range of a number system is the ratio of the largest value that can be represented with this system to the smallest one.} of data represented by low precision FXP is limited. This makes FXP suitable to represent data with only narrow range of values. Since this is not the case for most DNNs, using low precision FXP for DNNs is challenging. To enable this, various approaches were adopted such as quantization \cite{gholami2021survey}. For instance, uniform quantization includes scaling weights and activations of DNN and mapping them to a restricted range of values. These values can be represented by low-bit-width FXP. This allows lowering the number of bits to be less than 8 bits \cite{lo2020energy,Anwar2015Fixed,lin2016fixed}, and even as low as 2 bits (i.e., ternary DNNs \cite{hwang2014fixed, kim2021zero, mellempudi2017mixed}) or 1 bit (i.e., binary DNNs \cite{wang2019learning, rastegari2016xnor, samragh2021application, yin2020xnor, lin2015neural}). For more information about the FXP quantization, precision reduction, and binary DNNs the interested reader is referred to \cite{gholami2021survey, sze2017efficient, qin2020binary}.


In short, the FXP for DNN implementation offers great hardware efficiency at the expense of some accuracy degradation. Between the two extreme representations (FLP and FXP), there are several number systems that offer different trade-offs (Pareto optimal points) between the hardware efficiency and the acquired accuracy. These number systems and their usage for DNN implementation are presented in subsequent sections of this paper. 

\section{LNS for DNN Architectures}\label{LNS_Sec}
 
Proposals for LNS first emerged in the 1970s to implement the arithmetic operations of digital signal processing. The utilization of LNS for neural computing was first proposed in the late 90's \cite{arnold1997cost}. Since then, using LNS to implement efficient hardware for DNN has become more popular. The main benefit of using LNS is in simplifying the implementation of the costly arithmetic operations required for DNN inference and/or training \cite{parhami2020computing}. In addition, representing the data in LNS enables a reduction of the number of bits required to obtain the same DNN accuracy as with conventional number systems \cite{miyashita2016convolutional,sanyal2020neural}. In LNS, a real number $n$ is represented with a logarithm of radix $a$ of its absolute value ($\tilde{n}=\log_a(|n|)$) and a sign bit $s_n$\footnote{Some works use additional dedicated bit $z_n$ to indicate when $n$ equals zero \cite{alam2021low, kouretas2018logarithmic}, while others use a special code to represent zero \cite{parhami2020computing}.} \cite{parhami2020computing}. The number $\tilde{n}$ is represented using two's complement fixed point format \cite{alam2021low}, as shown in Figure \ref{lns_rep}. The radix $a$ of the logarithm is usually selected to be 2 for simpler hardware implementation. Throughout this survey, we will use $a=2$ as well.

The main DNN operation that can be dramatically simplified using LNS is the multiplication by transforming it into linear (i.e., fixed point) addition. The LNS product $\tilde{p}$ of two real numbers $n_1,\text{ and } n_2$ is calculated as follows 
\begin{flalign}\label{lns-mul}
\tilde{p}& =\tilde{n_1} \odot \tilde{n_2},&\\ \nonumber
     & =\log_2(|n_1| \times |n_2|),& \\ \nonumber
      & =\tilde{n_1} + \tilde{n_2}, \\ \nonumber
\end{flalign}
\begin{equation} \label{product_sign}
 s_{\tilde{p}}=s_{n_1} XOR \   s_{n_2}, 
 \end{equation}
where $\odot$ is the multiplication operation in LNS domain that can be implemented with a simple integer adder, and $s_{\tilde{p}}$ is the product sign, which is calculated by XORing the signs ($s_{n_1}\text{ and } s_{n_2}$) of the two operands. 

Existing proposals for LNS-based DNNs are for either using LNS for the whole DNN architecture from end-to-end, just for using the LNS-based multipliers, or for using logarithmic quantization for DNN weights and/or layer inputs. Based on this classification, LNS-based DNN architectures are discussed next by highlighting the challenges associated with each architecture and the solutions presented in the related work.

\subsection{End-to-end LNS-based DNN Architectures}
 End-to-end-LNS implementation utilizes the LNS for all blocks of the architecture, and thus, no conversion from or to conventional systems takes place. For this, the inputs (i.e., the dataset) and the weights\footnote{When the architecture targets DNN inference.} are assumed to be fed to DNN in LNS format. This task is usually performed offline and has no overhead on the implemented architecture. In this section, we review LNS-domain implementation of the main operations that are needed for DNN training and inference. The two types of DNNs that were implemented using LNS from end to end are convolutional neural networks (CNNs) \cite{miyashita2016convolutional, sanyal2020neural} and recurrent neural networks (RNNs) \cite{kouretas2018logarithmic}. These two types of DNN have different architectures, but they share the same basic operations which are multiplication, addition, and activation functions. Since the multiplication operation becomes a linear addition in LNS-domain, the challenging part of this architecture is implementing LNS-addition and LNS-activation functions, which are discussed next.
 
\subsubsection{Addition in LNS}

\begin{figure}[h]
    \centering
    \includegraphics[width=0.5\textwidth]{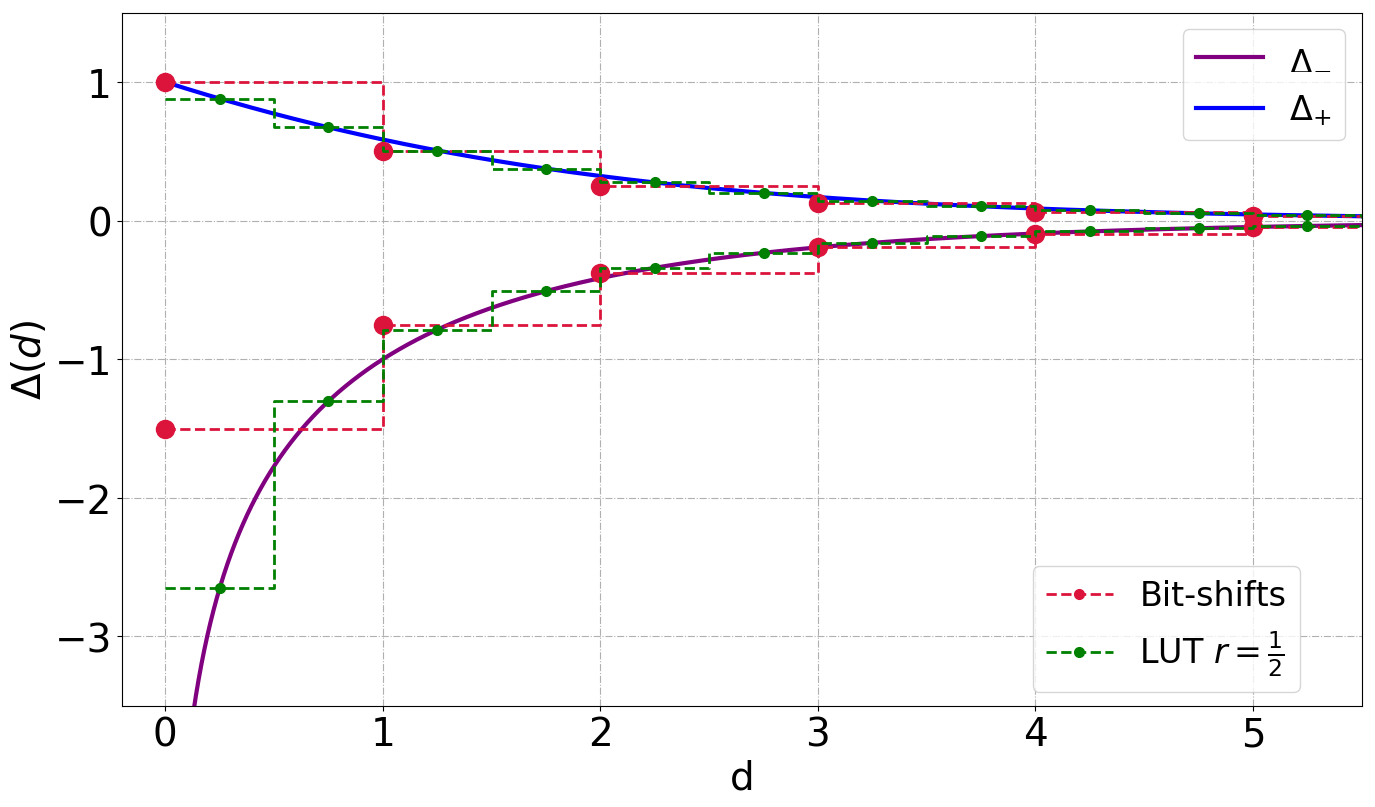}
    \caption{LUTs with $r=0.5$ and bit-shift approximations for the terms $\Delta_ {\pm}= \log_2(1\pm2^{-|\tilde{n_1}-\tilde{n_2}|})$ \cite{sanyal2020neural}}
    \label{LNS_add}
\end{figure}

As opposed to multiplication, performing addition in LNS is not straightforward. Let $\tilde{n_1}$ and $\tilde{n_2}$ be the two operands to be added in LNS. This LNS addition $\oplus$ is usually defined as follows

\begin{flalign}\label{lns-add}
\tilde{sum}& =\tilde{n_1} \oplus \tilde{n_2},&\\ \nonumber
     & =\log_2|(-1)^{s_{n_1}}\times2^{\tilde{n_1}} +(-1)^{s_{n_2}}\times 2^{\tilde{n_2}}|, \\ \nonumber
\end{flalign}
where $\tilde{sum}$ is the LNS domain summation of the two operands, and $s_{n_1} \text{ and } s_{n_2}$ are their signs. As these operands can be negative or Positive, $\tilde{sum}$ is derived from \ref{lns-add} \cite{sanyal2020neural} such that 
	\begin{align}\label{lns-add_2}
	\tilde{sum}=
	\left\{
	\begin{array}{ll}
	\max(\tilde{n_1},\tilde{n_2})+\log_2(1+2^{-|\tilde{n_1}-\tilde{n_2}|}),\\ \ \ \  \ \ \  \ \ \  \ \ \  \ \ \  \ \ \  \ \ \  \ \ \  \ \ \  \ \ \ \text{if  } s_{n_1} =s_{n_2},\\
	\max(\tilde{n_1},\tilde{n_2})+\log_2(1-2^{-|\tilde{n_1}-\tilde{n_2}|}), \\ \ \ \  \ \ \  \ \ \  \ \ \  \ \ \  \ \ \  \ \ \  \ \ \  \ \ \  \ \ \ \text{if  } s_{n_1} \neq s_{n_2}, 
	\end{array}
	\right.
	\end{align}

		\begin{align}\label{lns-add_2}
	s_{\tilde{sum}}=
	\left\{
	\begin{array}{ll}
	s_{n_1},& n_1>n_2,\\
	s_{n_2}, & n_1 \leq n_2, 
	\end{array}
	\right.
	\end{align}
	where $s_{\tilde{sum}}$ is the sign of the summation. To reduce the computational complexity of calculating $\tilde{sum}$, the term $\Delta_{\pm}= \log_2(1\pm2^{-|\tilde{n_1}-\tilde{n_2}|})$ is approximated using look-up-tables (LUTs) \cite{sanyal2020neural, kouretas2018logarithmic} or reduced to be implemented via bit-shifts \cite{sanyal2020neural, miyashita2016convolutional}. The LUT approximation requires using LUT of size $r_{max}/r$, where $r_{max}$ is the range of values stored in the LUT and $r$ is the resolution of the stored values. For the bit-shift implementation, the approximation in (\ref{log_approxi}) is utilized to replace the calculation of $\log_2(1\pm2^{-|\tilde{n_1}-\tilde{n_2}|})$ by a simple shift operation illustrated in (\ref{bs_approxi}).  
  \begin{equation} \label{log_approxi}
 \log_2 (1+x)=x, \text{ for } 0<x<1, 
 \end{equation}
   \begin{equation} \label{bs_approxi}
 \log_2(1\pm2^{-|\tilde{n_1}-\tilde{n_2}|})= \pm \mathbf{BS}(1,-|\tilde{n_1}-\tilde{n_2}|),
 \end{equation}
 where $\mathbf{BS}(b,d)=b \times 2^d$ means to shift the bits of $b$ binary representation by $|d|$ Positions, to the left if $d$ is negative and to the right otherwise. Figure \ref{LNS_add} shows how these two approximations are almost equivalent, However, LUTs require circuits with larger silicon and add extra delays to the system \cite{kouretas2018logarithmic}.

\subsubsection{Activation Functions in LNS}
Some activation functions can be transformed directly to the LNS domain by using the LNS operations to implement them. For example, the Leaky-ReLU for a number $n$ in linear domain, shown in (\ref{lrelu-linear}), is simply represented in LNS domain as in (\ref{lrelu-lns}) \cite{sanyal2020neural}.
	\begin{align}\label{lrelu-linear}
	LReLU(n|\alpha)=
	\left\{
	\begin{array}{ll}
	\alpha n, & n<0,\\
	n ,& n\geq0,
	\end{array}
	\right.
	\end{align}

	\begin{align}\label{lrelu-lns}
	\tilde{LReLU}(	(\tilde {n}, s_n)|\alpha)=
	\left\{
	\begin{array}{ll}
	(\tilde {n}+\alpha, s_a=s_n),& s_n=1,\\
	(\tilde {n}, s_a=s_n), & s_n=0, 
	\end{array}
	\right.
	\end{align}
	where $\alpha$ is a constant, $s_a$ is the sign after applying the activation, and $\tilde {n}$ and $s_n$ are the logarithm and the sign of $n$. 
However, for more complicated activation functions, such as Sigmoid, $\tanh$, and Softmax, more efficient hardware is obtained if these functions are approximated with piece-wise approximation that can be implemented using combinational logic \cite{kouretas2018logarithmic}. The motivation of this is that the approximation becomes an additional source of non-linearity and places a low burden on the performance of the implemented DNN architecture. This is particularly the case if these functions are used within the training process which is inherently noisy \cite{sanyal2020neural}. As an example, equation (\ref{sig-lns}) shows the LNS domain piece-wise approximation of $\tanh$ activation function in (\ref{sig})  \cite{kouretas2018logarithmic}. 

 \begin{align} \label{sig}
\tanh(n)=  \frac{1-e^{-n}}{1+e^{-n}}
\end{align}

	\begin{align}\label{sig-lns}
	\tilde{\tanh}(\tilde {n}, s_n)\approx 
	\left\{
	\begin{array}{llll}
	(0, s_a=s_n),& \tilde {n}> 0,\\
	(\tilde {n}, s_a=s_n), & -10<\tilde {n}\leq 0, \\
	(0,  s_a=1),& \tilde {n}<-10,\\
	(0,  s_a=1),& \tilde {n}=0, z_n=1,
	\end{array}
	\right.
\end{align}

\subsubsection{Summary and Discussion of End-to-End LNS-based DNN architectures}
When LNS is used to represent DNN data from end to end, all operations needed to perform DNN training and / or inference must be implemented in the LNS domain. Multiplication is implemented using FXP addition, while other operations, such as addition and activation functions, need to be approximated. The presented approximation techniques that have been proposed for DNN implementation introduce insignificant loss in the performance of  the implemented architectures. The classification accuracy degradation is found to be less than 1\% for the studied end to end LNS-based CNN architectures \cite{miyashita2016convolutional,sanyal2020neural}. These work do not investigate the associated impact on the CNN hardware efficiency. However, an idea about the hardware efficiency is offered by LNS implementation of long short-term memory (LSTM) architecture where the area is saved by 36\% for a 9-bit design \cite{kouretas2018logarithmic}, while the area savings decrease when the number of bits increased, due to the LUTs required for the LNS addition approximation.      

\subsection{LNS Multiplier-based DNN architectures}
Since end-to-end LNS implementations of DNNs introduces complexity for implementing additions and activation functions, an alternative approach is to limit using the LNS to the multipliers. The focus here is to design an efficient LNS-based Multiplier that receives linear operands and produces a linear product as well.  
  
\subsubsection{LNS-based Multiplier}

To simplify the discussion and comparisons between various LNS multipliers proposed for DNN, some notations are introduced for the next subsections. Let $n$ be a Positive integer and its $w$-bit binary representation is $B_n= b_{w-1}\ b_{w-2}\dots b_0$. Let $b_k$ be the most significant $'1'$ in $B_n$ ($k$ is called the characteristic number of $n$). The linear number $n$ and its logarithm can be represented by
\begin{equation} \label{number}
n=2^{k} (1+x), 
\end{equation}
\begin{equation} \label{log}
\log_2 (n)=k+\log_2 (1+x), 
\end{equation}
where $0\leq x < 1$ is called the mantissa of $n$. Let $n_1=2^{k_1} (1+x_1)$ and $n_2=2^{k_2} (1+x_2)$ be the multiplier and multiplicand, respectively. The product of these operands and its logarithm are given by
\begin{equation} \label{prod}
n_1 \times n_2=2^{k_1+k_2} (1+x_1) (1+x_2), 
\end{equation}
\begin{equation} \label{prod_log}
\log_2(n_1 \times n_2)=k_1+k_2+\log_2 (1+x_1)+\log_2 (1+x_2), 
\end{equation}

The main idea of the logarithmic multiplier is to use a specific approximation based on the characteristics of the logarithms to simplify the product calculation by mainly using shift and add operations instead of hardware-intensive conventional multiplication. Given their effectiveness, many logarithmic multipliers have been proposed for image processing and neural computing \cite{saadat2018minimally, mitchell1962computer, babic2011iterative,ansari2019hardware, pilipovic2020design,harsha2022low}. Several of these multipliers were utilized to also build efficient DNN architectures. They are classified in this survey into multipliers that use Mitchell's approximation, iterative logarithmic multipliers, double-sided error multipliers, and multipliers with explicit logarithm and antilogarithm modules.    

\paragraph{Mitchell's Multiplier}\label{Mitchell_sec}
 According to Mitchell's algorithm \cite{mitchell1962computer}, the logarithm of a number $n$ is approximated with piece-wise straight lines as in (\ref{log_approxi}). Thus, the logarithm of the product in (\ref{prod_log}) is approximated by the sum of the characteristic numbers and the mantissas of the operands as follows
 \begin{equation} \label{prod_app}
\log_2(n_1 \times n_2)\approx k_1+k_2+x_1+x_2.  
\end{equation}

The final product is obtained in (\ref{ma_prod}) by applying the antilogarithm on (\ref{prod_app}) using the approximation in (\ref{log_approxi}). Then, the product of two integers is calculated using add and shift operations, as

	\begin{align}\label{ma_prod}
	n_1 \times n_2 \approx
	\left\{
	\begin{array}{ll}
	2^{k_1+k_2} (1+x_1+x_2), & x_1+x_2 < 1\\
	2^{k_1+k_2+1} (x_1+x_2), & x_1+x_2 	\geq 1 
	\end{array}
	\right.
	\end{align}

Even though the error introduced by Mitchell's approximation is relatively high (up to 11\% \cite{babic2011iterative}), this multiplier showed no accuracy degradation for CNN architecture with 32- bit precision \cite{kim2018low}, while being 26.8\% more power-efficient compared to conventional multipliers of the same number of bits. 

To gain additional power efficiency over the one achieved by Mitchell's multiplier, a truncated-operand approach has been proposed \cite{kim2018efficient}. Instead of using the whole operands, these operands are truncated and only their $\omega$ most significant bits are used to calculate the approximated product. For instance, selecting $\omega=8$ allows for a more efficient multiplier that saves up to 88\% and 56\% of power when compared to an exact 32-bit FXP multiplier and a Mitchell's multiplier, respectively. The additional error introduced by this truncation caused an accuracy degradation of 0.2\% for the ImageNet dataset. The significant power saving associated with the negligible performance degradation of this approach comes from the fact that the most significant part of the operand can be sufficient to provide an acceptable approximation \cite{sarwar2016multiplier, hashemi2015drum}.

\begin{figure}[h]
    \centering
    \includegraphics[trim={1.8cm 0cm 0cm 0},clip,width=\linewidth]{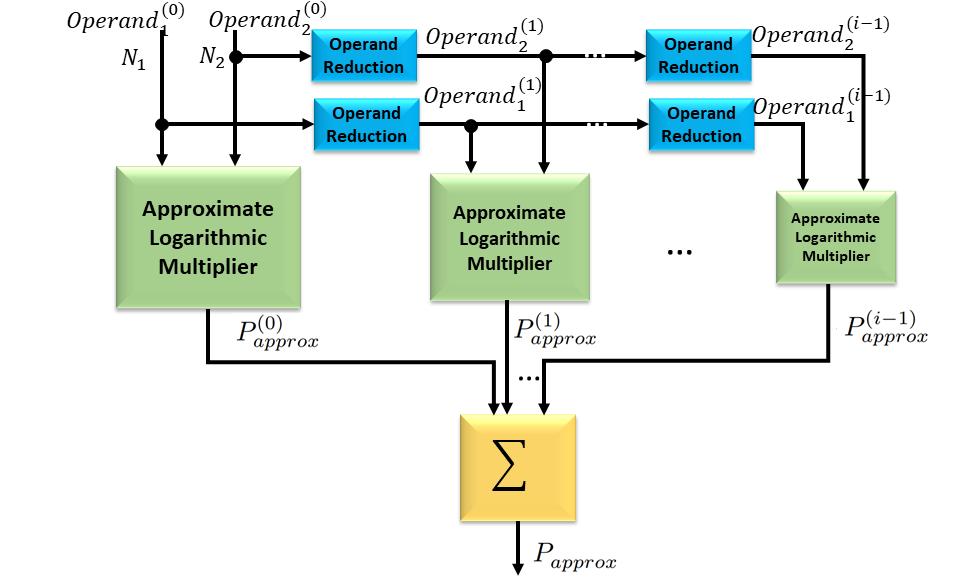}
    \caption{Iterative logarithmic multiplier}
    \label{iterative}
\end{figure}

\paragraph{Iterative Logarithmic Multiplier}
The iterative logarithmic multiplier aims to reduce the error introduced by the approximation in (\ref{ma_prod}) by adding correction terms. The calculation of these terms usually requires iterative multiplications that can be calculated in the same way as calculating the approximate product ( see Figure \ref{iterative}). These correction terms can be biased (always Positive) or unbiased (negative/Positive).

The product of two numbers in (\ref{prod}) can be written using biased correction terms \cite{babic2011iterative} as

\begin{flalign}\label{babic_prod}
n_1 \times n_2  =P_{approx}+ E,\\ \nonumber
\end{flalign}

where $P_{approx} =2^{k_1+k_2}(1+x_1+x_2)$ is an approximate product that can be calculated using shift and add operations. $E=2^{k_1+k_2} x_1 x_2$ is a correction term that is ignored in (\ref{ma_prod}). Estimating the term $E$ requires calculating the product $(2^{k_1} x_1) (2^{k_2} x_2) = (n_1-2^{k1}) (n_2-2^{k2})$ iteratively, in the same way of calculating $P_{approx}$. Then, 
\begin{flalign}
n_1 \times n_2 & =P_{approx}^{(0)}&+& E^{(0)}, &\\ \nonumber
     & =P_{approx}^{(0)}&+& P_{approx}^{(1)}+E^{(1)}, \\ \nonumber
          & =P_{approx}^{(0)}&+& P_{approx}^{(1)}+\dots +P_{approx}^{(i-1)}+E^{(i-1)},  \nonumber
\end{flalign}
where $i$ is the number of iterations and $E^{(i-1)}$ is the error to be ignored after the $i^{th}$ iteration. Notice that when $i$ equals the number of bits that have the value of '1' in the operands, then $E^{(i-1)} = 0$, and the exact product is produced. For each iteration, the new operands to be multiplied are obtained by removing the leading 'ones' from the original operands. For this reason, the correction terms can be calculated in parallel using one additional circuit for each iteration. Hence, there is a trade-off between the accuracy of the multiplication and the area and power overhead due to adding these correction circuits. For example, this iterative logarithmic multiplier with one iteration (i.e., one correction circuit) was able to save 10\% on area and 20\% on power consumption without any notable impact on the learning accuracy when it was used to implement the hardware of a relatively simple neural network and compared with the case of using floating point multiplier \cite{lotrivc2011logarithmic}. 

On the other hand, using unbiased iterative correction terms of (\ref{kim19}) shows a better area and power reduction by up to 44.6\% and 48.1\%, respectively, compared to the multiplier designed with the error terms of (\ref{babic_prod}) \cite{kim2019cost}.

\begin{align}\label{kim19}
	E =
	\left\{
	\begin{array}{ll}
	((1-x_1)2^{k_1}-1) ((1-x_2)2^{k_2}-1), \\ \qquad\qquad \qquad \text{ if   } x_1+x_2+2^{-k_{1,2} }\geq 1\\
	(x_1 2^{k_1}) (x_2 2^{k_2}), \\ \qquad \qquad\qquad  \text{ if   } x_1+x_2+2^{-k_{1,2} }< 1, 
	\end{array}
	\right.
	\end{align}
where $k_{1,2}=max(k_1, k_2)$. Using 32-bit multipliers based on the correction terms in (\ref{kim19}) and (\ref{babic_prod}) gives classification accuracy comparable to that of default floating point multiplier for various CNN architectures \cite{kim2019cost}.
	
\paragraph{Double-sided Error Multiplier}

The logarithmic approximation in (\ref{log_approxi}) underestimates the logarithm value and results in an always negative error. Since tolerating and even having better performance in the presence of noise is an important feature of DNNs, creating a multiplier with "double-sided error" can enhance the implemented architecture of DNN in terms of accuracy and hardware efficiency \cite{ansari2019improving}. To achieve this, another logarithmic approximation may be utilized \cite{ansari2020improved}. 

In addition to the expression in (\ref{number}), any integer number $n$ can be represented as  
\begin{equation} \label{number2}
n=2^{k+1} (1-y), 
\end{equation}
where $0\leq y < 1$. The two representations in (\ref{number}) and (\ref{number2}) can be used to come up with a new logarithmic approximation with double-sided error \cite{ansari2020improved}, as 
	\begin{align}\label{ans_prod}
	\log_2 (n) \approx
	\left\{
	\begin{array}{ll}
	k+x, & n-2^k <2^{k+1}-n.\\
	k+1-y, & \text{otherwise.} 
	\end{array}
	\right.
	\end{align}
	
To utilize this approximation, the two multiplied operands, $n_1,\text{and }  n_2$, are transformed into the closest powers of two plus an additional negative or Positive term ($a_1,\  a_2$, respectively). Hence their product can be calculated as 
\begin{flalign}\label{ans_operands_pro}
n_1 \times n_2 & =(2^{k_1}+a_1) \times (2^{k_2}+a_2), &\\ \nonumber
     & =2^{k_1+k_2}+a_1 2^{k_2}+a_2 2^{k_1}+a_1a_2.  \nonumber
\end{flalign}

\begin{figure}[h]
    \centering
    \includegraphics[width=0.5\textwidth]{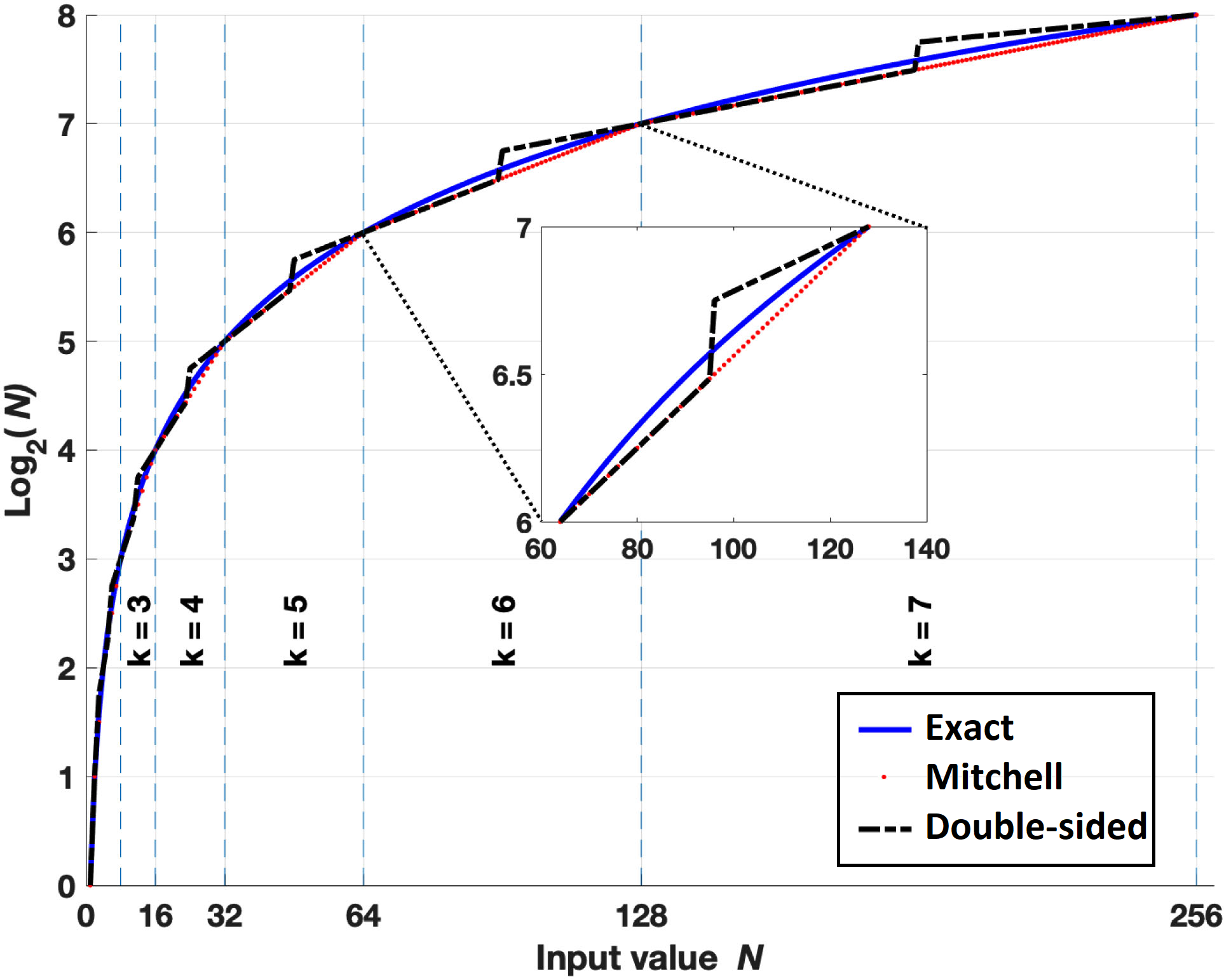}
    \caption{Mitchell's vs. double-sided approximations of $\log_2 N$ \cite{ansari2020improved}}
    \label{ansari_mitchels}
\end{figure}

The product is approximated to be the sum of the first three terms in (\ref{ans_operands_pro}), whereas the last term ($a_1a_2$) is omitted as an approximation error. In fact, this approximation has a larger absolute error compared to Mitchell's approximation (of (\ref{log_approxi})). This can be observed as well from Figure \ref{ansari_mitchels}. However, the signed errors help with canceling the error and having higher classification accuracy by up to 1.4\% compared to the case of using a conventional exact multiplier to implement CNNs \cite{ansari2020improved}. This comes in addition to the better hardware efficiency indicated by 21.85\% of power savings.

\paragraph{Explicit Logarithm-Antilogarithm Multiplier}
For the aforementioned multipliers (i.e., Mitchell's, iterative,.. ) there is no explicit module for logarithm and antilogarithm calculation. The implementation of these operations is not done explicitly, but their characteristics are used to transform the costly multiplication into simpler operations. On the other side, the logarithmic multiplier can be designed by explicitly transforming the operands into the logarithmic domain, adding the operands, and finally returning back to the linear domain.
As calculating the exact logarithm is very costly, the logarithm/antilogarithm operations are usually approximated using LUTs \cite{johnson2018rethinking} or bit-level manipulation \cite{juang2018area}.

One approach that uses LUT-based approximated log/antilog multipliers for CNN is presented in \cite{johnson2018rethinking}. 
Let $n$, represented in (\ref{number}), be an operand in the linear domain. The logarithm of this operand is approximated as  
\begin{flalign}\label{johnson_log}
\log_2 (n)&=k+\log_2 (1+x),&\\ \nonumber
     & \approx k+Q_{\beta}(\log_2 (Q_{\gamma}(1+x))),  \nonumber
\end{flalign}	
where $Q_{\gamma}$ is the quantization used to represent $(1+x)$ with $\gamma$ bits in the linear domain, and $\beta$ bits are guaranteed for the approximated logarithm in the log domain using $Q_{\beta}$. The mapping from $1+x \text{ into } \log_2 (1+x)$ is obtained using a LUT of size $2^\gamma \beta$ bits. If the product in LNS domain is represented by $\tilde{p}= m.f$, where $f \in  [0, 1)$ is the fraction part and $m$ is the integer one, the approximated antilog of this product is calculated by
\begin{flalign}\label{johnson_anti}
\log_2^{-1}(\tilde{p}) & =2^{m}\ 2^{f}, &\\ \nonumber
     & \approx2^{m} \ (1+Q_{\alpha}(2^f-1)),  \nonumber
\end{flalign}	
i.e., the term $2^{m}$ is implemented by a bit shift, whereas $2^{f}$ is approximated using a LUT. The LUT maps $f \text{ to } Q_{\alpha}(2^f-1)$, where $Q_{\alpha}$ is the applied quantization to limit the number of bits to $\alpha$.  

LUT-approximation is usually used to escape from the errors introduced by the approximation in (\ref{log_approxi}). However, to keep the size of the required LUTs reasonable, the value of $\alpha, \gamma, \text{ and } \beta$ should be kept as small as possible. This introduces a loss in accuracy. In addition, this approach is expected to be less hardware-friendly because of the needed overhead to implement these LUTs. Nevertheless, experimental results showed that integrating 16-bit LUT-based multiplier with a wide FXP accumulator results in a reduction in power consumption and area by up to 59\% and 68\%, respectively, in comparison to 16-bit FLP multiplier \cite{johnson2018rethinking}. This comes in addition to achieving a negligible accuracy degradation ($<1\%$) for the CNN ResNet50 network trained on the ImageNet dataset.

Another approach for approximating log/antilog modules is using bit-level manipulation to innovate area and speed efficient logarithm or antilogarithm operations  \cite{juang2018area,juang2016lower}. Among these works, the two-region manipulation-based logarithm converter and bit correction-based antilogarithm converter \cite{juang2011high} are used to implement an LNS multiplier that is exploited to build an efficient CNN accelerator design from area and delay point of view \cite{juang2018area}. When this design is compared to conventional multiplier implementation, it saves up to 60\% of the area-delay product. However, neither the accuracy of the CNN nor a comparison with other logarithmic multipliers has been reported for this design.

\begin{table*}[t]\fontsize{8pt}{13}\selectfont
  \begin{threeparttable}[b]

	\caption{Comparison of DNN architectures based on logarithmic multipliers. This table considers the targeted machine learning (ML) phase (inference/training ), DNN type, the number system of the implemented arithmetic, the precision of the numbers indicated by number of bits, the dataset used to test the approach, and the tested DNN models. Comparison metrics (accuracy loss, power  and area savings) are extracted from the related reference. Mostly, the enhancement or loss is reported in comparison to FLP32, when applicable. When the results is compared to another number system, this is indicated by mentioning this number system between Parentheses.}
	\centering
	\label{lns_multipliers_comp}
\begin{tabular}{|c|c|c|ccc|c|c|c|c|c|c|}
\hline
\multirow{2}{*}{Ref.} & \multirow{2}{*}{ML Phase} & \multirow{2}{*}{DNN}                               & \multicolumn{3}{c|}{DNN Arithmatic}                                                                                                                                                      & \multirow{2}{*}{bits} & \multirow{2}{*}{\begin{tabular}[c]{@{}c@{}}Tested\\ Dataset\end{tabular}} & \multirow{2}{*}{\begin{tabular}[c]{@{}c@{}}Tested\\ model\end{tabular}}            & \multirow{2}{*}{\begin{tabular}[c]{@{}c@{}}Accuracy\\ loss \%\end{tabular}} & \multirow{2}{*}{\begin{tabular}[c]{@{}c@{}}Power\\ Saving \%\end{tabular}} & \multirow{2}{*}{\begin{tabular}[c]{@{}c@{}}Area  \\ Saving \%\end{tabular}} \\ \cline{4-6}
                      &                           &                                                    & \multicolumn{1}{c|}{Multiplier}                                                           & \multicolumn{1}{c|}{Adder}                                                      & Activation &                         &                                                                           &                                                                                    &                                                                             &                                                                            &                                                                             \\ \hline
\cite{kim2018low }               & Inference                 & CNN                                                & \multicolumn{1}{c|}{Mitchell’s}                                                            & \multicolumn{1}{c|}{Linear}                                                     & Linear     & 32                      & \begin{tabular}[c]{@{}c@{}}MNIST,\\ CIFAR-10\end{tabular}                 & \begin{tabular}[c]{@{}c@{}}LeNet,\\  Cudaonvnet\end{tabular}                       & \begin{tabular}[c]{@{}c@{}}0\end{tabular}                        & \begin{tabular}[c]{@{}c@{}}26.8\end{tabular}                    & -                                                                           \\ \hline                      

\cite{kim2018efficient}               & Inference                 & CNN                                                & \multicolumn{1}{c|}{\begin{tabular}[c]{@{}c@{}}Truncated\\ Mitchell's\end{tabular}}       & \multicolumn{1}{c|}{Linear}                                                     & Linear     & 32                      & \begin{tabular}[c]{@{}c@{}}MNIST,\\ CIFAR-10,\\ ImageNet\end{tabular}     & \begin{tabular}[c]{@{}c@{}}LeNet,\\  cuda-convnet,\\    AlexNet\end{tabular}       & \begin{tabular}[c]{@{}c@{}}0.2\\ (FXP)\end{tabular}      & \begin{tabular}[c]{@{}c@{}}96\\ (FXP)\end{tabular}                       & \begin{tabular}[c]{@{}c@{}}81\\ (FXP)\end{tabular}                        \\ \hline
\cite{lotrivc2011logarithmic}            & Inference                 & NN                                                 & \multicolumn{1}{c|}{Iterative}                                                            & \multicolumn{1}{c|}{Linear}                                                     & Linear     & 18                      & Proben1                                                                   & Custom                                                                             & -                                                                           & \begin{tabular}[c]{@{}c@{}}20 \end{tabular}                     & \begin{tabular}[c]{@{}c@{}}10\end{tabular}                       \\ \hline
\cite{kim2019cost}               & Inference                 & CNN                                                & \multicolumn{1}{c|}{Iterative}                                                            & \multicolumn{1}{c|}{Linear}                                                     & Linear     & 32                      & \begin{tabular}[c]{@{}c@{}}CIFAR-10,\\ ImageNet\end{tabular}              & \begin{tabular}[c]{@{}c@{}}NiN,\\ AlexNet,\\  GoogLeNet,\\ ResNet-50\end{tabular} & \begin{tabular}[c]{@{}c@{}}\textless{}0.5\end{tabular}           & -                                                                          & -                                                                           \\ \hline
\cite{ansari2020improved}            & Inference                 & \begin{tabular}[c]{@{}c@{}}DNN,\\ CNN\end{tabular} & \multicolumn{1}{c|}{\begin{tabular}[c]{@{}c@{}}Double-sided\\ error\end{tabular}}         & \multicolumn{1}{c|}{\begin{tabular}[c]{@{}c@{}}Linear\end{tabular}} & Linear     & 8                       & \begin{tabular}[c]{@{}c@{}}MNIST,\\ CIFAR-10\end{tabular}                 & \begin{tabular}[c]{@{}c@{}}Custom,\\ Alexnet\end{tabular}                          & \begin{tabular}[c]{@{}c@{}} -1.4\tnote{1}\end{tabular}              & \begin{tabular}[c]{@{}c@{}}21.85\end{tabular}                   & -                                                                           \\ \hline
\cite{juang2018area}             & Inference                 & CNN                                                & \multicolumn{1}{c|}{\begin{tabular}[c]{@{}c@{}}Explicit\\ Log/Antilog\end{tabular}} & \multicolumn{1}{c|}{Linear}                                                     & Linear     & 32                      & -                                                                         & Custom                                                                             & -                                                                           & -                                                                          & \begin{tabular}[c]{@{}c@{}}$\sim$76\end{tabular}                 \\ \hline

\cite{johnson2018rethinking}           & Inference                 & CNN                                                & \multicolumn{1}{c|}{\begin{tabular}[c]{@{}c@{}}Explicit\\ Log/Antilog\end{tabular}}       & \multicolumn{1}{c|}{\begin{tabular}[c]{@{}c@{}}Linear\end{tabular}} & Linear     & 16                      & ImageNet                                                                  & ResNet-50                                                                          & \begin{tabular}[c]{@{}c@{}}\textless{}1\end{tabular}             & \begin{tabular}[c]{@{}c@{}}59\\ (FLP16)\end{tabular}                      & \begin{tabular}[c]{@{}c@{}}68\\ (FLP16)\end{tabular}                       \\ \hline

\end{tabular}
     \begin{tablenotes}
       \item [1] Accuracy enhancement

     \end{tablenotes}
       \end{threeparttable}
\end{table*}

\paragraph{Summary and Discussion of LNS-based Multipliers}
LNS-based multipliers use the characteristics of the logarithm to transform the multiplication into simpler operations. Most of the proposed logarithmic multipliers for DNN architecture started from Mitchell's approximation to innovate their logarithmic multipliers. Table \ref{lns_multipliers_comp} compares various architectures that use LNS multipliers. We notice that these multipliers are used to implement efficient CNN architectures suitable for DNN inference rather than training. In addition, this table depicts that using the vanilla Mitchell's multiplier offers power-efficient hardware with comparable accuracy to the FLP32 multiplier of the same number of bits. However, reducing the number of bits requires a more accurate approximation with less average error than Mitchell's. When the LNS multiplier is designed with the characteristics of DNN in mind, such as the double-sided-error multiplier, the outcome is further reduction in the number of bits with significant power savings, while preserving and even enhancing the classification accuracy of the reported CNNs.    

\subsection{Logarithmic quantization for DNN architectures}
\begin{figure*}[h]
    \centering
    \includegraphics[width=1\textwidth]{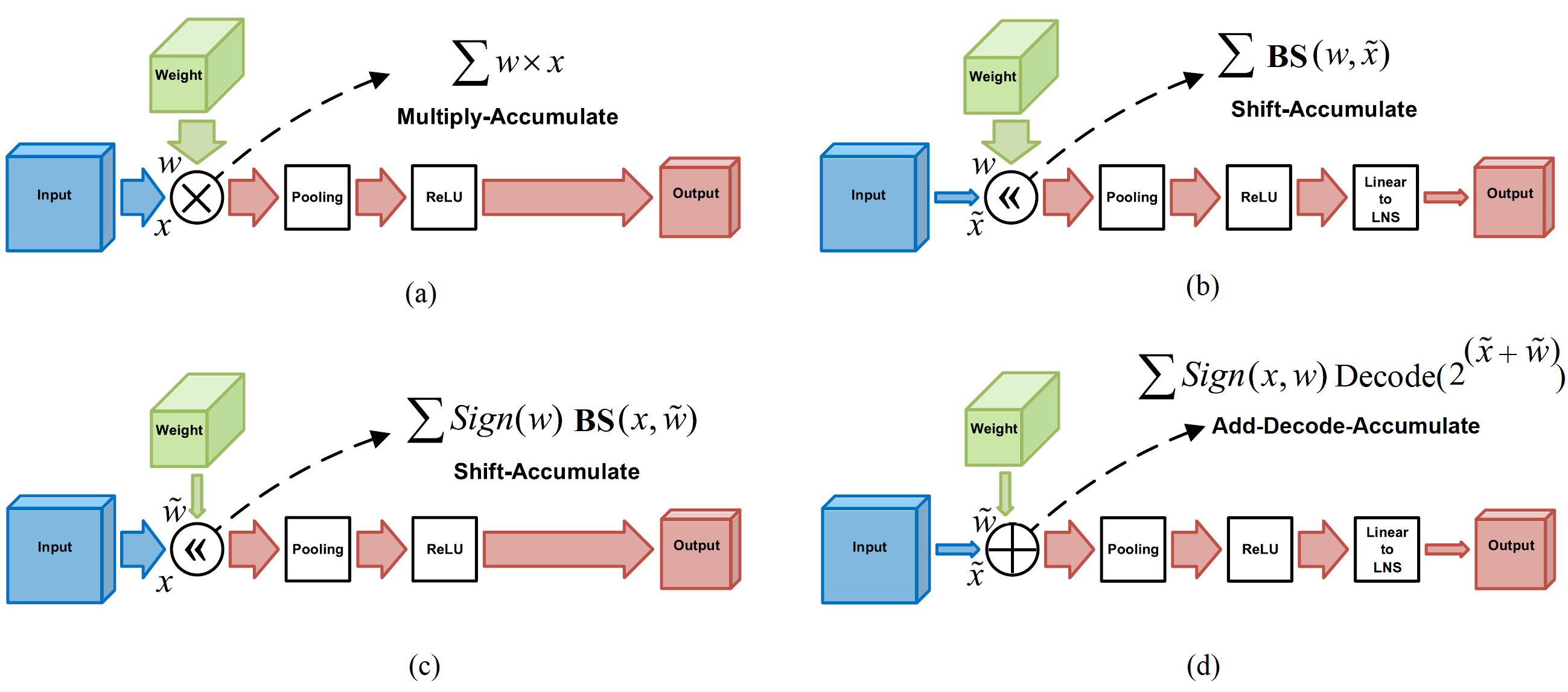}
    \caption{Arithmetic processing elements of CNNs utilizing linear and various logarithmic quantization schemes (a) Linear quantization (b) Inputs logarithmic quantization (c) Weights logarithmic quantization (d) Inputs and weights logarithmic quantization, adapted from \cite{xu2020base}}
    \label{LNS_Quantize}
      
\end{figure*}

Logarithmic quantization involves representing a real number $n$ with a sign and an integer exponent (integer power of two). The integer is usually an approximation of the logarithm $\log_2|n|$ of the real number after applying the clipping and rounding \cite{zhao2021low, vogel2018efficient}. Logarithmic quantization has been employed in order to achieve efficient hardware implementation of CNNs \cite{zhao2021low, vogel2018efficient}. The main idea behind this is the that the multiplication by this integer exponent can be easily implemented in hardware by bit shifting. In CNNs, both the convolutional and fully-connected layers include matrix multiplication, i.e., dot product between the weights $w$ of each layer and the input activation $x$, which is the output of the previous layer after applying the non-linearity (e.g., ReLU). This matrix multiplication is usually performed using a number of multiply-and-accumulate operations when the conventional data representation is used to implement digital hardware, as shown in Figure \ref{LNS_Quantize}(a). However, this dot product can be implemented more efficiently when Logarithmic quantization is utilized. Due to the non-uniform distribution of the weights and inputs, 
using nonuniform quantization, such as logarithmic quantization, is preferred over the uniform quantization, such as when FXP is used \cite{vogel2018efficient}.

Existing CNN architectures that use logarithmic quantization assume that weights and/or the inputs of the layer are quantized. When logarithmic quantization is applied to inputs only (i.e activations), Figure \ref{LNS_Quantize}(b), or to weights only, Figure \ref{LNS_Quantize}(c), the dot product becomes a simple bit shift operation followed by an accumulation. applying logarithmic quantization on the weights only shows insignificant accuracy degradation \cite{lu2020very,lee2017lognet} and significant power and area savings \cite{xu2018low, xu2020base}, see Table \ref{lns_quant_comp}. Quantizing the inputs only in LNS results in the same performance from an accuracy point of view \cite{ueki2018learning, miyashita2016convolutional}, however, with an additional linear-to-LNS module to be added. This module is responsible for transforming output activations to LNS before storing them in memory. This scheme has the advantage of requiring a smaller memory bandwidth as the stored activation is represented in LNS \cite{miyashita2016convolutional}. 

The works that apply logarithmic quantization to weights as well as to activations usually use a logarithm radix different from '2' \cite{vogel2018efficient, zhao2021low}. Then, the multiplication becomes an addition of the LNS quantized weights and activations followed by an approximation to decode this sum into the linear domain before implementing the accumulation. This add-decode-accumulate scheme adds a complication to hardware implementation, however, with comparable accuracy to the aforementioned logarithmic quantization schemes, as illustrated in Table \ref{lns_quant_comp}.

\begin{table*}[t]\fontsize{7pt}{12}\selectfont
  \begin{threeparttable}[b]
	\caption{Comparison of DNN architectures based on logarithmic quantization}
	\centering
	\label{lns_quant_comp}
\begin{tabular}{|c|c|c|ccc|c|c|c|c|c|c|}
\hline
\multirow{2}{*}{Ref.} & \multirow{2}{*}{ML phase}                                       & \multirow{2}{*}{DNN} & \multicolumn{3}{c|}{CNN Arithmetic Operations}                             & \multirow{2}{*}{bits} & \multirow{2}{*}{\begin{tabular}[c]{@{}c@{}}Tested  \\ Dataset\end{tabular}} & \multirow{2}{*}{\begin{tabular}[c]{@{}c@{}}Tested\\    Model\end{tabular}}           & \multirow{2}{*}{\begin{tabular}[c]{@{}c@{}}Accuracy\\ loss \%\end{tabular}} & \multirow{2}{*}{\begin{tabular}[c]{@{}c@{}}Power  \\ Saving \%\end{tabular}} & \multirow{2}{*}{\begin{tabular}[c]{@{}c@{}}Area  \\ Saving \%\end{tabular}} \\ \cline{4-6}
                      &                                                                 &                      & \multicolumn{1}{c|}{Multiplier} & \multicolumn{1}{c|}{Adder}  & Activation &                         &                                                                                  &                                                                                      &                                                                             &                                                                              &                                                                             \\ \hline
\cite{xu2020base}                & Inference                                                       & CNN                  & \multicolumn{1}{c|}{Shift}      & \multicolumn{1}{c|}{linear} & Linear     & 5                       & ImageNet                                                                         & \begin{tabular}[c]{@{}c@{}}AlexNet,\\ VGG16,\\  ResNet34,\\ DenseNet161\end{tabular} & 2.5                                                                     & \begin{tabular}[c]{@{}c@{}}55.60\\ (FXP)\end{tabular}                      & \begin{tabular}[c]{@{}c@{}}50.42\\ (FXP)\end{tabular}                     \\ \hline
\cite{xu2018low}                & Inference                                                       & CNN                  & \multicolumn{1}{c|}{Shift}      & \multicolumn{1}{c|}{Linear} & Linear     & 4                       & ImageNet                                                                         & AlexNet                                                                              & \begin{tabular}[c]{@{}c@{}} -3.9 \tnote{1}\end{tabular}                        & \begin{tabular}[c]{@{}c@{}}73.74 \\ (FXP16)\end{tabular}                   & \begin{tabular}[c]{@{}c@{}}58.03   \\ (FXP16)\end{tabular}                \\ \hline
\cite{lu2020very}                & \begin{tabular}[c]{@{}c@{}}Inference\\ \end{tabular}       & CNN                  & \multicolumn{1}{c|}{Shift}      & \multicolumn{1}{c|}{Linear} & Linear     & 4\&8                    & ImageNet                                                                         & \begin{tabular}[c]{@{}c@{}}AlexNet,\\ VGG16,\\ ResNet-18/34, \\ YOLOv2\end{tabular}  & \textless{}1                                                              & -                                                                            & -                                                                           \\ \hline
\cite{lee2017lognet}               & Inference                                                       & CNN                  & \multicolumn{1}{c|}{Shift}      & \multicolumn{1}{c|}{Linear} & Linear     & 4                       & ImageNet                                                                         & \begin{tabular}[c]{@{}c@{}}VGG16,\\  AlexNet\end{tabular}                            & \textless{}1                                                              & -                                                                            & -                                                                           \\ \hline
\cite{ueki2018learning}              & \begin{tabular}[c]{@{}c@{}}Training \end{tabular} & DNN                  & \multicolumn{1}{c|}{Shift}      & \multicolumn{1}{c|}{Linear} & Linear     & 7                       & MNIST                                                                            & Custom                                                                               & \begin{tabular}[c]{@{}c@{}}1.6  \end{tabular}                      & \begin{tabular}[c]{@{}c@{}}-90 \\ (FXP32)\end{tabular}                     & \begin{tabular}[c]{@{}c@{}}290   \\ (FXP32)\end{tabular}                  \\ \hline
\cite{miyashita2016convolutional}         & \begin{tabular}[c]{@{}c@{}}Training\end{tabular} & CNN                  & \multicolumn{1}{c|}{Shift}      & \multicolumn{1}{c|}{Linear} & Linear     & 5                       & CIFAR10                                                                          & \begin{tabular}[c]{@{}c@{}}AlexNet, \\ VGG16\end{tabular}                            & \textless{}1                                                              & -                                                                            & -                                                                           \\ \hline
\cite{vogel2018efficient}             & Inference                                                       & CNN                  & \multicolumn{1}{c|}{Add-decode} & \multicolumn{1}{c|}{Linear} & Linear     & 8                       & \begin{tabular}[c]{@{}c@{}}CIFAR-10,\\  ImageNet, \\ Cityscapes\end{tabular}     & \begin{tabular}[c]{@{}c@{}}AllCNN, \\ VGG16, \\ Dilated\end{tabular}                 & \begin{tabular}[c]{@{}c@{}}\textless{}3\end{tabular}              & \begin{tabular}[c]{@{}c@{}}22.3\\  (FXP)\end{tabular}                      & \begin{tabular}[c]{@{}c@{}}35  \\ (LUTs)\end{tabular}                     \\ \hline
\cite{zhao2021low}              & Training                                                        & CNN                  & \multicolumn{1}{c|}{Add-decode} & \multicolumn{1}{c|}{Linear} & Linear     & 8                       & \begin{tabular}[c]{@{}c@{}}CIFAR-10, \\ ImageNet, \\ SQuAD, \\ GLUE\end{tabular} & \begin{tabular}[c]{@{}c@{}}ResNet, \\ BERT\end{tabular}                              & \textless{}0.3                                                           & 90                                                                         & -                                                                           \\ \hline
\end{tabular}
     \begin{tablenotes}
       \item [1] Accuracy enhancement

     \end{tablenotes}
       \end{threeparttable}

\end{table*}

\section{RNS for DNN Architectures}\label{RNS_Sec}

\begin{figure}[!ht]
\centering
\includegraphics[scale=0.33]{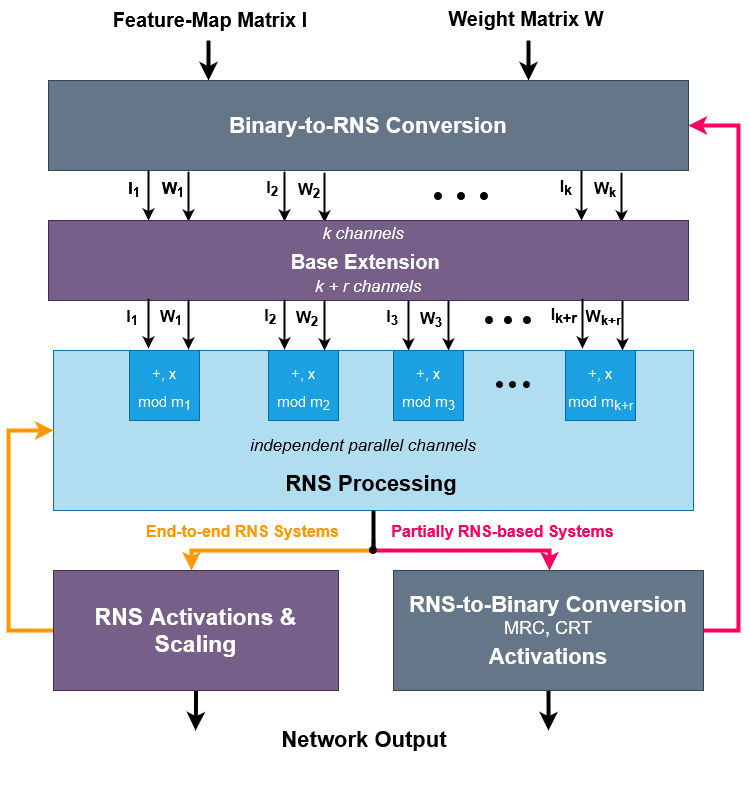}
\caption{ Computation flow of RNS accelerators. Partially RNS-based accelerators utilize binary converters between successive layers for non-trivial RNS operations, while end-to-end systems perform all NN operations, including activation functions, in the RNS domain. Base-extension is usually required to increase the dynamic range before the accumulation of the partial products.}
\label{f:rnsproc}
\end{figure}

The Residue Number System (RNS) can be an attractive choice for DNN accelerators due to its arithmetic properties. In this Section, a brief overview of the RNS is given, and several RNS-based architectures for AI applications reported in literature are presented. Architectures are classified to \textit{partially RNS-based}, where intermediate conversions to conventional representations between successive layers are used, and \textit{end-to-end} RNS-based architectures, where the entire processing takes place in the RNS domain. The typical computation flow of these two types of systems is shown in Fig.~~\ref{f:rnsproc}.

The number representation scheme utilized in realizing DNN architectures directly impacts the accuracy, speed, area, and energy dissipation. Modern Deep Learning models keep growing in depth and number of parameters and require a huge amount of elementary arithmetic operations, the majority of which are multiply-add operations (MAC). 

In the Residue Number System, each number is represented as a tuple of residues with respect to a modulus set  $\{m_1,m_2,\ldots,m_n\}$, which is called the \textit{base} of the representation. The dynamic range of the representation is given by 
\begin{equation}
   R =  \prod_{i=1}^{N} m_i .
    \label{eq:drange}
\end{equation}
If the moduli are \textit{co-prime}, i.e., 
\begin{align}
    \gcd_{\substack{1\leq i,j\leq N\\ i\neq j}} (m_i,m_j ) = 1,
\end{align}
where $\gcd(\cdot)$ denotes the greatest common divisor operation, each integer inside the range $[0, R)$ has a unique RNS representation 
\begin{equation}
     X \mapsto(x_1,x_2, \ldots, x_n) , \ x_i = \langle X \rangle_{m_i},
     \label{eq:forw}
\end{equation} where $\langle \cdot \rangle _{m}$ is the modulo-$m$ operator. Inverse transformation is generally harder and can be realized by means of the \textit{Mixed Radix Conversion} or the \textit{Chinese Remainder Theorem} \cite{rnsbook}. 

\subsection{RNS Addition and Multiplication}
Due to the properties of the modulo operation, addition and multiplication can be done independently and in parallel for each residue channel, i.e., without inter-channel propagation of information. Suppose $ A = (a_1,a_2,\ldots,a_n) $ and  $ B = (b_1,b_2,\ldots,b_n) $, then
\begin{equation}
   a \oplus b = ( \langle a_1 \oplus b_1\rangle_{m_1}, \langle a_2 \oplus b_2\rangle_{m_2} , \ldots , \langle a_n \oplus b_n\rangle_{m_n} ),
\end{equation}
where $\oplus$ can be either the addition or the multiplication operator. This property is what makes RNS very efficient in applications that require a large number of these operations, such as DSP applications and, more recently, Neural Network inference. This is because, by decomposing the computations into independent channels, long carry propagation chains are eliminated, thus arithmetic circuits can operate at higher frequencies, or with reduced power dissipation.

\begin{figure}[!ht]
\centering
\includegraphics[scale=0.33]{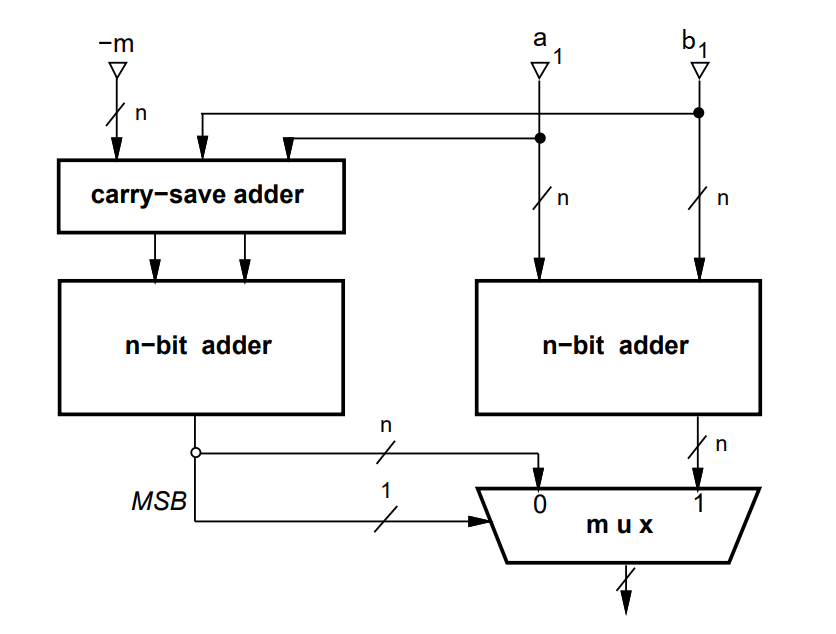}
\caption{ Modulo adder \cite{survey}}
\label{f:modadd}
\end{figure}

The general architecture of a modulo adder is shown in Fig~\ref{f:modadd} \cite{survey}. The design consists of an \mbox{$n$-bit}, adder, where n is the size of the channel, that performs the addition of two numbers $a + b$, and a CSA adder which performs the computation of $a + b - m_i$ (modulo operation). The sign of the CSA result is used to select the correct result of the two adders.

The selection of moduli can significantly simplify the design of modulo arithmetic circuits. In case of moduli of the form $2^k$ the modulo operation translates into just keeping the $k$ least significant bits, whereas in the case of $2^k-1$, the output carry of the addition simply needs to be added to the result. In this case, end-around-carry adders can be used. For channels of the form $2^k+1$, diminished-1 arithmetic can be used \cite{dim1}, which basically involves an inverted end-around logic. If the size of the channel is large, then fast adder designs such as prefix adders must be utilized within each channel.

Modulo multiplication is a trickier operation, however the benefits of RNS can be greater. This is because of the (approximately) quadratic scaling of a multiplier with the input size. This means that, by decomposing a large multiplication into smaller ones, the energy and delay savings can be significant, providing that the overhead of the modulo is diminished.  One approach for RNS multiplication is to perform regular multiplication of the two $n$-bit numbers and then use a reduction circuit to obtain the final result modulo $m_i$. This approach introduces, however, considerable overhead to the design, as the reduction of a $2n$-bit number to a $n$-bit number modulo $m_i$ is not as straightforward as in the case of addition. A low complexity adder-based combinatorial multiplier has been proposed in \cite{combmult}, where the number of FAs required is minimized.  Other multiplication techniques are based on intermediate RNS transformations, such as \textit{core functions}~\cite{core} and \textit{isomorphisms}~\cite{survey}, which are transformations that convert multiplication into addition. These transformations utilize look-up tables to convert RNS to an intermediate representation where multiplication is translated into addition.

\begin{figure}[!ht]
\centering
\includegraphics[scale=0.22]{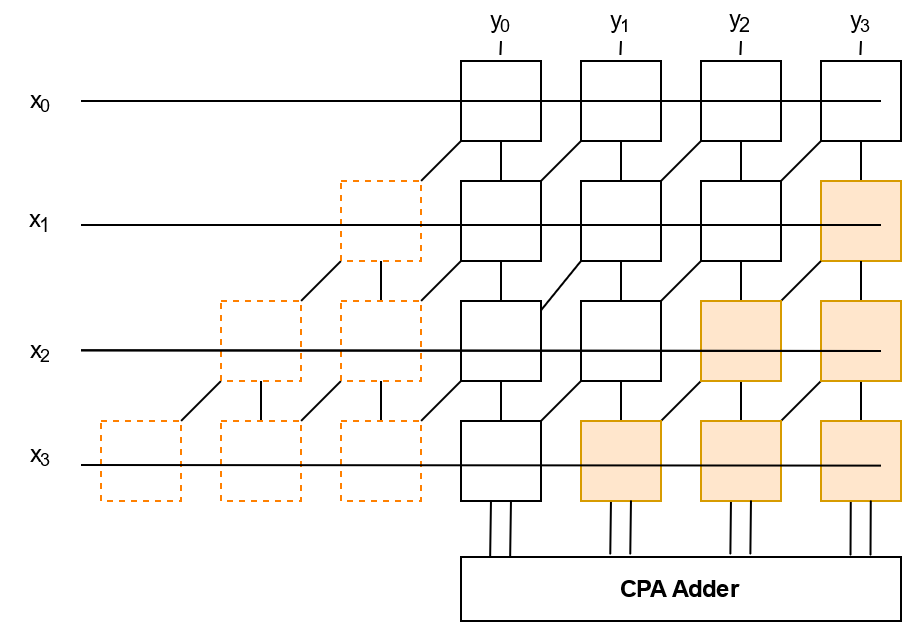}
\caption{ Array multiplier for calculating  $x\cdot y \bmod 15$. At each level $i$, the full adders corresponding to the $i$ most significant positions (dashed squares) are moved to the $i$ least significant positions. This is possible because $2^{n+k} \bmod (2^n - 1) = 2^k$. A modulo-15 carry-propagation adder is used to obtain the final result.} 
\label{f:modmult}
\end{figure}

In the case of modulo~$2^k$ multiplication, regular multipliers operating only on the $k$ LSBs can be used, whereas in the case of modulo-$(2^k + 1)$ diminished-1 arithmetic can be applied \cite{dim1mult}. A end-around-carry multiplier which can be used for $2^k-1$ channels is shown in Fig.~\ref{f:modmult}. Due to  the properties of the particular channel, the modulo operation is translated into simple bit re-ordering, thus no overhead is introduced. 

Based on the above, most of the RNS designs reported in literature utilize thes low-cost forms of moduli, which allow to fully exploit the RNS benefits (elimination of long carry chains) with minimal hardware overhead. 

\subsubsection{Conversions and Non-trivial Operations}

While addition and multiplication are very efficiently implemented in RNS, other operations such as sign detection, comparison and division, or the realization of non-linear activation functions are not straightforward to implement, as the require the combination of the RNS channels.  A common approach is to use RNS-to-binary converters and then perform the operation in the binary domain.  
Conversion to and from an RNS representation is a crucial for the performance of any RNS-based processing system. Especially for the architectures that perform frequent intermediate conversions (partially RNS-based) the overhead can be significant. The complexity of these converters largely depends on the particular base selection, namely the size, number and format of the moduli. 

While  \textit{Binary-to-RNS} or \textit{forward} converters can have a relative simple hardware realization, following Eq.~\ref{eq:forw}, especially if particular forms of moduli are used, \textit{RNS-to-binary} or \textit{inverse} converters are generally harder to implement. Extensive bibliography exists for this topic. The most commonly used approaches are the Chinese Remainder Theorem (CRT) and the Mixed Radix Conversion (MRC) \cite{mrc}. The CRT is expressed as
\begin{equation}
    X = \left \langle \bigg (\sum_{i=1}^n \overline{m}_i \langle x_i\overline{m}_i^{-1} \rangle _{m_i}\bigg)\right \rangle_M 
    \label{eq:crt}
\end{equation}
 where $\langle \cdot \rangle$ denotes the modulo operation, $X$ is the binary representation of the number, $x_i$ are its residues, $m_i$ are the moduli, $M$ is the dynamic range, $\overline{m}_i = M/m_i $, and $\overline{m}_i^{-1} $ is the modulo inverse of $\overline{m}_i$. CRT requires the pre-computation of $\overline{m}_i$, and $\overline{m}_i^{-1} $, additions of potentially large products, as well as the final modulo operation with the $M$, which can be very large.  It can be computed, however, in a single cycle. In the other hand, MRC requires the computations of some intermediate coefficients and is a sequential process which requires several steps, but these steps only include small bit-width operations.
The Mixed Radix Conversion finds the coefficients $ {k_1, k_2, \ldots, k_n}$, such that 
\begin{equation}
    X = k_1 + k_2m_1 + k_3m_1m_2 + \ldots + k_nm_1m_2\ldots m_{n-1}
    \label{eq:mrc}
\end{equation}

The coefficients are calculated one by one in a number of steps \cite{mrc}, each of which requires the previously calculated coefficients. The modulo inverses can be pre-calculated and pipelining stages can be introduced to make this computation efficient. 

Sign detection is one of the most critical and frequent operation required by NNs, as the Rectified Linear Unit (ReLU), which maps negative values to zero, is the most common activation function. In an RNS representation, numbers in the range $0 < X <R/2$ are positive, whereas numbers in the range $R/2 \leq x < M$ are negative.  Magnitude comparison of two RNS numbers which is required for the MaxPooling layers, is also difficult to directly to implement in the RNS domain. Comparison algorithms for particular moduli sets ($2^k-1,2^k,2^k+1$) ~\cite{comparison}, or more complex general ones have been proposed~\cite{comp}, that can eliminate the overhead of the conversion. If the choice of moduli is restricted to some specific bases, simple and efficient algorithms have been reported for sign detection~\cite{sign} and comparison. Finally, division, which is necessary after the multiplication and accumulation operations of a convolutional layer for example, in order to bring the result in the original dynamic range, also requires special handling. Methods that use special form of moduli, such as powers
of two~\cite{scale2} or a product of the moduli~\cite{scaling} as divisors can simplify the hardware implementation. Some methods rely on using small (only one-channel wide) lookup-tables and typically relay on base extension methods, during which an RNS base with $k$ channels is extended to $k+r$ channels.

\subsection{Partially RNS-based Architectures}

A common approach in RNS-based DNN implementations is to perform all multiply-add operations of a single convolutional or dense layer in the RNS representation and then use a converter to obtain a partial result in normal positional binary representation \cite{matmul,nestedrns,rnscnn,rnsinfer}. With this intermediate result, the non-linear activation functions (\textit{ReLU}, $\tanh$, \textit{softmax}) can be computed and the results can be again converted to RNS format to be fed to the next layer.
\par Many application-specific AI accelerator designs, as well as more general purpose architectures, such as TPUs or GPUs, perform DNN computations by decomposing them into matrix or vector multiplication primitives. Thus, by utilizing efficient hardware matrix multipliers, performance can be orders of magnitude better than CPUs. An RNS TPU (Tensor Processing Unit) is proposed in \cite{matmul}. In the core of this architecture there is a RNS matrix multiplier implemented as a two dimensional systolic array. Each processing element performs one operation (MAC) at each cycle, and passes the result to neighboring processing elements. Systolic arrays are an efficient way of increasing throughput and dealing with the limited memory bandwidth problem. In this particular RNS systolic array, each processing element decomposes the larger MAC operation (typically 8 or 16 bits), into smaller, each within the range of the respective channel, that can be performed in parallel. Using an FPGA implementation the RNS matrix multiplier is reported to perform a $32\times32$ fixed point matrix multiplication up to $9\times$ more efficiently than a binary matrix multiplier for large matrices.

In \cite{rnscnn} the authors extend the RNS usage to the implementation of the convolution operation. Individual layers are executed on an RNS-based FPGA accelerator. However results are sent to a CPU, which performs the non-trivial RNS operations, such as applying the activation functions, before being sent back to the FPGA for the execution of the next layer. RNS results in a  \mbox{$7.86\%$ -- $37.78\%$}  reduction of the hardware costs of a single convolutional layer compared to the two’s-complement implementation, depending on the RNS base selection.

A variant of the Residue Number System, called the \textit{Nested RNS} (NRNS) is proposed in \cite{nestedrns}. NRNS applies a recursive decomposition of the residue channels into smaller ones.
Adder and multipliers  can be thus implemented by using smaller and faster circuits.
Assuming that a number $X$ has a RNS representation of $(x_1, x_2,\ldots, x_n)$, then the nested RNS representation will be of the form 
\begin{equation}
    X = (x_1,x_2,\ldots,(x_{i1},x_{i2},\ldots,x_{im}),\ldots,x_n)
\label{nrns}    
\end{equation}
where $(x_{i1},x_{i2},\ldots,x_{im})$ is the RNS decomposition of the $i-th$ channel
 This technique introduces an additional complexity, as any operation must be recursively applied to each level of the representation, however it manages to handle large dynamic ranges with very small channels. The authors use a 48-bit equivalent dynamic range composed only of 4-bit MAC units which can be realized by look-up tables of the FPGA. Contrary to \cite{rnscnn}, which relies on an external CPU, in this work binary-to NRNS and NRNS-to-binary conversions are realized by DSP blocks and on-chip BRAMs. After Input data are converted into the NRNS representation, a number of parallel convolutional units perform all the necessary computations of a single convolutional layer. The results are then  converted to binary using a tree-based NRNS-to-binary converter. The authors report a performance per area improvement of $5.86\times$  compared to state-of-the-art FPGA implementations for the ImageNet benchmark. In a different approach  the RNS arithmetic costs are reduced by restricting the RNS base selection to low-cost moduli of the form $2^k\pm 1$ \cite{rnsinfer}. This way, modified fast prefix adders and CSA trees using end-around-carry propagation can be used, diminishing any overhead of the modulo operator.
 
 In another category of RNS-based architectures, the usage of very small channels allows the realization of multiplier-free CNN architectures. The authors utilize a small RNS base of (3,4,5) and reduce the implementation of the multiplications to shifts and additions \cite{residuenet}. Despite the reduced dynamic range of the representation, the authors report minimal accuracy loss, while achieving $36\%$ and $23\%$ reduction in power and area, respectively. A method to drastically reduce the number of multiplications in CNN RNS-bases accelerators is proposed in \cite{icecsSak2021}. It utilizes a modified hardware mapping of the convolution algorithm where the order of operations is rearranged. Because of the small dynamic range of each RNS channel, there is an increased  number of common factors
inside the weight kernels during convolution. By first executing the additions of the input feature map terms that correspond to the same factors, and then performing the multiplications with the common weight factors,  a $97\%$, reduction of the total multiplications is reported for state-of-the-art CNN models. 

\subsection{End-to-end RNS Architectures}

While the above circuits mange to achieve some performance gain in the implementation of a single convolutional layer, they require significant amounts of extra hardware to perform the conversions which can become the bottleneck for some of these designs. More recent approaches focus on overcoming the difficulties of performing operations such as sign detection, comparison, and scaling which is usually required following multiplication. In these approaches, input data are initially converted to an RNS representation and then the entire processing takes place in the RNS domain.

\subsubsection{State-of-the-art End-to-end RNS Architectures}

 The system in \cite{rnsdnn} introduces some novel mechanisms for dealing with this problem and proposes an efficient fully RNS-based architecture.
The authors of this work choose to work with moduli of the form $2^k-1$, $2^k$, $2^{k+1}-1$. In particular they select (31,32,63) as the basis of their representation, as it is found to provide a sufficient dynamic range (16-bit equivalent), that results in no accuracy loss, for state-of-the-art networks and benchmarks. For the design of the modulo adders, which are simplified due to the particular selection of the moduli, parallel-prefix Sklansky adders with an end around carry are utilized. For the multiplications, a radix-4 Booth encoding is adopted within each channel. An optimized sign detection unit for this set of moduli is used, based on an approach proposed in \cite{sign}.
which can be further transformed and result in  a relatively hardware-friendly implementation.
Using a similar logic to the work proposed in \cite{comparison}, the comparison of two RNS numbers can also be implemented by calculating auxiliary partitioning functions.

The authors also introduce a base extension mechanism which is necessary in order to avoid potential overflow when accumulating the partial sums. In this work, a base extension method proposed in \cite{be} is used, where the middle channel is extended from $2^k$ to $2^{k+e}$. This way the convenient properties of the chosen moduli are maintained. Base extension takes places once before each multiplication to ensure that the product lies within the dynamic range and then again before the accumulation. The authors define the number of extra bits that are added each time based on extensive simulation on benchmark networks and on a per-layer basis. RNS circuits result in significant delay and energy efficiency improvement, especially in the case of multiplication at the cost of larger overall area. Comparisons in terms of various performance metrics against  the Eyeriss \cite{eyeriss} accelerator are reported for various networks. Up to 61\% reduction in energy consumption compared to the conventional positional binary representation has been achieved. The system can also support an increased clock frequency as high as $1.20$ GHz versus $667$ MHz in the case of the positional binary system, indicating a 1.8$\times$ improvement in  computational latency.

\begin{table*}[t]\fontsize{7pt}{12}\selectfont
\caption{RNS-based architectures targeting DNN applications.\label{t:rndnn}}
\begin{tabular}{|l|c|c|c|c|c|c|c|c|c|}
\hline
Ref                        & Application & \begin{tabular}[c]{@{}c@{}}RNS Arithmetic \\ Details\end{tabular} & RNS Base                     & Dataset                                                   & NN Model                                                                    & Platform                                                 & \begin{tabular}[c]{@{}c@{}}Accuracy\\ drop (\%) \end{tabular} & Speedup ($\times$)   & \begin{tabular}[c]{@{}c@{}}Energy\\ reduction (\%)\end{tabular}                                                   \\ \hline
\cite{nestedrns}    & CNN                  & Partialy - Nested                                                          & no specific                           & ImageNet                                                           & ConvNet2                                                                    & FPGA                                                     & -                                                       & 3.1   & -                                                          \\ \hline
\cite{residuenet}   & CNN                  & Partially - Mult. free                                                     & \{3,4,5\}                             & \begin{tabular}[c]{@{}c@{}}MNIST, CIFAR10,\\ ImageNet\end{tabular} & \begin{tabular}[c]{@{}c@{}}LeNet, VGG, \\ AlexNet, \\ Resnet50\end{tabular} & FPGA                                                     & 0.03                                                 & 2.8   & 36                                                      \\ \hline
\cite{rnsdnn}       & CNN                  & End-to-end                                                                 & \{31,32,63\}                          & ImageNet                                                           & \begin{tabular}[c]{@{}c@{}}AlexNet, VGG-16,\\ SqueezeNet\}\end{tabular}     & ASIC                                                     & 0.27                                                 & 2.9   & 30-61                                                    \\ \hline
\cite{rnsmem}       & CNN                  & \begin{tabular}[c]{@{}c@{}}End-to-end \\ (in-memory)  \end{tabular}                                                    & \{$2^k-1,2^k,2^k+1$\} & \begin{tabular}[c]{@{}c@{}}MNIST, INDOOR, \\ CIFAR10\end{tabular}  & custom                                                                      & \begin{tabular}[c]{@{}c@{}}Memristor \\ IMC\end{tabular} & 1                                                    & 1.6-5 & 500 - 840     \\ \hline
\cite{rnssim}       & CNN                  & \begin{tabular}[c]{@{}c@{}}End-to-end \\ (in-memory)  \end{tabular}                                               & \{31,32,63\}                          & \begin{tabular}[c]{@{}c@{}}MNIST,SVHN,\\ CIFAR10\end{tabular}      & DoReFa-Net                                                                  & \begin{tabular}[c]{@{}c@{}}DRAM \\ IMC\end{tabular}      & 1                         & 2-4.3 &90 -  100                \\ \hline
\cite{iscasSak2022} & LSTM                 & End-to-end                                                                 & \{3,5,7,31,32\}                       & Q-traffic                                                          & custom                                                                      & ASIC                                                     & 3                                                     & $1.6$   & 30                                                       \\ \hline
\end{tabular}
\end{table*}

\subsubsection{In-Memory Computing RNS Architectures}

Recently, there has been a growing focus of AI accelerator design research on in-memory computing. This is because of the paradigm-shifting effect that emerging memory technologies can have on processing systems. It is known that the largest part of the energy consumption of any DNN accelerator is due to the memory accesses and data transferring, particularly to and from the off-chip RAM. In-memory computing (IMC) aims to diminish data transfer costs by bringing the computing inside or near the memory elements.

Efforts have been made to bring the benefits of the RNS to IMC systems. In these (mainly digital) IMC designs, the benefit of the usage of RNS over binary representation stems from the speedup of the bitwise serial addition operations, due to the inherent parallel operations of the RNS channels.

 RNS has been utilized in the design of an in-memory computing system \cite{rnsmem}. In this work, the selected moduli are of the form  $2^k-1$, $2^k$, $2^k+1$. A sign detection mechanism similar to \cite{sign}, is developed, in order to implement the ReLU and MaxPooling operations without having to convert to a binary representation. Addition and multiplication within each RNS channel, take place inside the memory elements. Multiplication of two numbers, $a,b$ is implemented through addition and memory accesses by calculating the quantity $\frac{(a+b)^2}{4} - \frac{(a-b)^2}{4}$, where squaring is implemented using look-up-tables.
A single crossbar memory is assigned to each neuron and supports in-memory addition in a tree-based structure. For this purpose, a  Memristor Aided loGIC (MAGIC) is used. Based on experimental results, the proposed RNS in-memory architecture consumes 145.5$\times$ less energy and leads to a speedup of $35.4 \times$  compared to NVIDIA GPU GTX 1080. 

An near-memory RNS-based processing architecture is proposed in \cite{rnssim}. Instead of memristor-based memory macros, a DRAM computational sub-array is utilized for the implementation of the MAC operations in the RNS domain, combined with parallel-prefix adders, to implement bitwise multiplication and accumulation. Unlike \cite{rnsdnn}, where multiplication is directly implemented in memory (by mapping to additions and squaring), here they are implemented by combining elementary bit-wise operations (AND, OR, XOR) between the operands. The authors also design a more flexible activation function unit which is based on a Mixed-Radix conversion. Similar to \cite{rnsdnn}, an RNS base of $(2^k-1$, $2^k$, $2^{k+1}-1)$ is utilized. Gains in the order of $331-897\times$  in terms of energy efficiency compared to GPU platforms are reported, and $2\times$ compared to other IMC designs.

\subsection{Summary of RNS-based DNN architectures}

RNS-based architectures targeting DNN applications are summarized in Table~\ref{t:rndnn}. The majority of these approaches utilize low cost moduli of the form $2^k-1$, $2^k$, $2^k+1$ to reduce the overhead of the modulo operator and are targeting CNNs. Most of these RNS accelerators can achieve speedups in the order of $1.5-3\times$ and can also me more energy efficient. IMC RNS-based systems exhibit the largest energy savings. Among conventional systems, \cite{rnsdnn} illustrates more clearly the applicability of the RNS in DNN architectures by proposing a fully RNS system which outperforms the binary state-of-the-art counterpart. The RNS usage is also extended to LSTM networks, by designing hardware friendly RNS activation units for the implementation of $tanh$ and $sigmoid$ functions \cite{iscasSak2022}.

In conclusion, the Residue Number System (RNS) can be an attractive number representation choice for DNN accelerators, and several RNS-based architectures have been reported recently targeting AI applications, due to its various advantages. RNS exhibits inherent parallelism at the residue channel processing level. It utilizes parallel computations along separate residue channels, where operations in each of them are performed modulo a specific modulus, with no need for information (carry or other) to be shared between residue channels.
 
The main challenge in designing an efficient RNS-based accelerator is to minimize or, possibly, eliminate the overhead introduced due to the implementation of the non-linear operations. Another key factor is the optimization of the moduli selection and the corresponding arithmetic circuits, to meet the accuracy requirements.

Some of the RNS systems proposed in recent literature only perform  the multiply-add (MAC) or matrix multiplication operation, required by the convolutional layers, in the RNS representation, and use intermediate converters between number systems for the non-linear operations.
More recently, completely RNS-based approaches have been proposed that eliminate the overhead introduced by these intermediate conversions to and from a traditional positional binary representation.

\section{BFP for DNN Architectures}\label{BFP}

BFP representation offers a middle ground between FLP and FXP formats. This representation is proposed to preserve accuracy comparable to full precision FLP and hardware efficiency comparable to FXP. This is achieved by representing numbers with an exponent and a mantissa similar to FLP to guarantee a wide dynamic range. However, instead of representing each value separately, a group (called here a block) of values has a common exponent while maintaining private mantissas. Let $N$ be a tensor that represent a block of $t$ elements initially represented in FLP as 
\begin{flalign}\label{tensor}
N & =(n_1, \dots n_i, \dots n_{t}), &\\ \nonumber
     & =((-1)^{s_1}\  m_1\  2^{e_1}, \dots (-1)^{s_i}\  m_i\  2^{e_i}, \dots (-1)^{s_{t}}\  m_{t}\  2^{e_{t}}). \\ \nonumber
\end{flalign}
This block is represented with BFP format as $\grave{N}$ such that
\begin{flalign}\label{tensor_B}
\grave{N} & =(\grave{n_1}, \dots \grave{n_i}, \dots \grave{n_{t}}), &\\ \nonumber
     & =((-1)^{s_1}\  \grave{m_1}, \dots (-1)^{s_i}\  \grave{m_i}, \dots (-1)^{s_{t}}\  \grave{m_{t}})\times 2^{\epsilon_{N}}, \\ \nonumber
\end{flalign}
where $\epsilon_{N}$ is a shared exponent between the elements of block $N$, and $\grave{m_i}$ is the aligned mantissa of element $i$ such that $\grave{m_i} = \mathbf{BS}(m_i,e_i-\epsilon_{N})$, where $\mathbf{BS}$ is the bit-shift operation. For large difference between the Private and shared exponents ($e_i-\epsilon_{N}$), this shifting causes some of the least-significant bits of the mantissa to be truncated. The truncation happens frequently when there are many outliers in a block, which in turn depends on the size of the block and the way the shared exponent is selected.

Since the dot product is the basic operation involved in DNN inference and training, the main target of BFP is to simplify the complex hardware required to perform this operation when FLP is used. For two blocks $\grave{N_1}$ and $\grave{N_2}$ represented in BFP, the dot product is calculated as 

\begin{flalign}\label{tensor_dot}
\grave{N_1}.{\grave{N_2}}^T & =((-1)^{s_{1,1}}\  \grave{m_{1,1}}, \dots (-1)^{s_{t,1}}\  \grave{m_{t,1}})\times 2^{\epsilon_{N_1}}. \\\nonumber &((-1)^{s_{1,2}}\  \grave{m_{1,2}}, \dots(-1)^{s_{t,2}}\  \grave{m_{t,2}})^T\times 2^{\epsilon_{N_2}}\\ \nonumber
     & =2^{\epsilon_{N_1}+\epsilon_{N_2}} \sum_{i=1}^{t} (-1)^{s_{i,12}} \grave{m_{i,1}}\times\grave{m_{i,2}},\\ \nonumber
\end{flalign}

where $s_{i,j}$ and $m_{i,j}$ are the sign and mantissa of the $i^{th}$ element in the $j^{th}$ block, respectively, $\epsilon_{N_j}$ is the shared exponent of the $j^{th}$ block, $s_{i,12}$ results from XORing $s_{i,1}$ and $s_{i,2}$, and $T$ stands for transportation. Equation (\ref{tensor_dot}) shows that the dot product of two blocks of size $t$ represented in BFP involves $t$ FXP multiplications of mantissas, $t-1$ FXP additions of the products, and one addition of the two shared exponents. The additional overhead compared to FXP representation comes from the hardware required to handle the shared exponent which mainly depends on the number of the blocks \cite{darvish2020pushing}. As a result, the performance of DNN in presence of BFP representation is determined by block partition scheme, shared exponent selection, and the bit-width of the mantissa and shared exponent, which will be discussed next.

\subsection{BFP Block Design}

\begin{figure}[h]
    \centering
    \includegraphics[trim={0 0cm 0cm 4cm},clip,width=\linewidth]{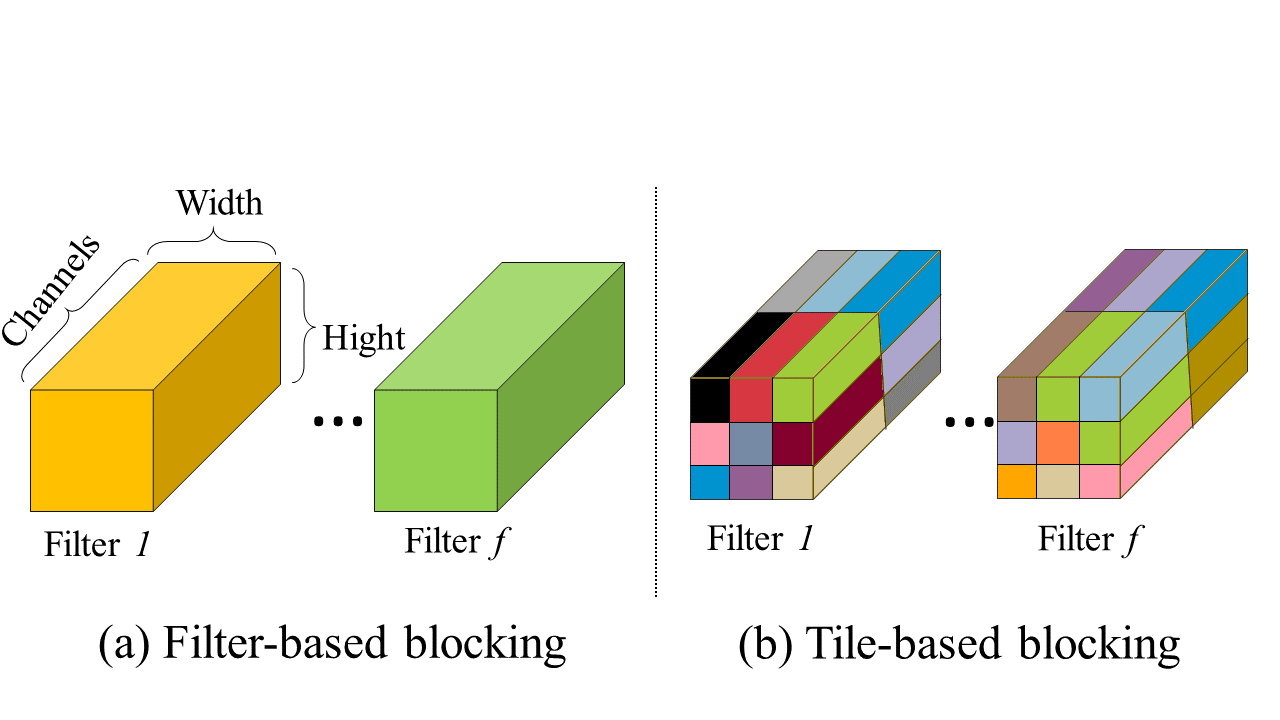}
    \caption{Different blocking Schemes, different colors indicate differed blocks}
    \label{blocking}
\end{figure}

Determining how the blocks are partitioned is essential to achieving good DNN performance with BFP \cite{lian2019high,song2018computation}. Usually, the input activation of each layer is considered as one block, whereas the weight matrix needs a specific scheme to be divided into blocks. There are two known blocking approaches, filter-based blocking \cite{lian2019high, song2018computation, yang2019swalp, fan2019static, ni2020lbfp, zhang2019block, koster2017flexpoint} and tile-based blocking \cite{zhao2021low, fan2018reconfigurable, zhang2021fast, darvish2020pushing, drumond2018training,wong2021low}, illustrated in Figure \ref{blocking}(a) and Figure \ref{blocking}(b), respectively. In the filter-based blocking, each filter of weights along the input channels is considered a block. Then the total number of blocks equals to the number of filters. This blocking is usually called coarse-grain blocking and it is the most hardware-friendly blocking approach as the accumulation of each output activation is done with the same shared exponent. Thus, it can be done using the FXP arithmetic \cite{lian2019high}. However, this approach may end up with severe accuracy degradation due to the increased number of outliers that need to be truncated within these large blocks.

On the other hand, the tile-based blocking is proposed to strike a compromise between accuracy and hardware efficiency. This approach relies on breaking large matrices of the filters down into small tiles to fit into limited hardware resources. Each tile is considered as a block with a shared exponent. The size of these tiles is a metric that need to be optimized. For example, a large tile of size 576 is used in \cite{drumond2018training} which requires a 12-bit mantissa to obtain acceptable accuracy. However, the authors in \cite{zhang2021fast} showed that 12-bit FXP can achieve similar accuracy with simpler hardware implementation. This indicates that BFP may has no advantage over FXP for such large tiles. Smaller tiles of 16 elements are used in \cite{darvish2020pushing, zhao2021low} seeking better accuracy, but with an added hardware complication comes from the need to convert to FLP before the accumulation.

\subsection{Shared Exponent Selection}
One shared exponent for each block need to be selected after partitioning the blocks\footnote{As the case of partitioning the weight blocks prior to DNN inference, which is usually performed offline.} and whenever a new block is created with multiple shared exponents. For example, this exponent is aligned after performing the calculation of each DNN layer as the calculation of the output activation usually ends up with a matrix of multiple exponents \cite{song2018computation}. To this end, most of DNN accelerators that adopt BFP calculate the shared exponent dynamically during DNN training or inference. Static shared exponent selection can be  utilized prior to DNN inference. 

One of two schemes is usually used for this dynamic shared exponent selection; maximal exponent-based or statistics-based schemes. The dynamic maximal exponent selection scheme is more popular \cite{aydonat2017opencl,jo2018training,das2018mixed}. In this scheme, for each block of (\ref{tensor}), different floating point numbers $n_i$ are compared and the maximum exponent is selected as follows
 \begin{equation} \label{shared_exp}
\epsilon_{N}=\max_{e^i}: i \in 1, \dots, t.  
\end{equation}
To find this maximal exponent before performing the dot product between weights and activations result from previous layer, the output activations represented in BFP with several exponents need to be converted back into FLP, which adds large overhead on the performance and the resources. 

To keep the advantage of the dynamic calculation of the shared exponent while avoiding frequent conversion between BFP and FLP the statistics-based scheme is proposed to predict the shared exponent during DNN training \cite{koster2017flexpoint, su2020processing}. In this scheme, the optimal exponent for each block is predicted based on statistics collected in the previous learning iteration. For example, in \cite{koster2017flexpoint} the maximum value recorded within each block is stored for the last $i$ iterations. Then, the maximum and the standard deviation of the stored values are used to calculate the shared exponent for the next iteration. This scheme works because the values within each block change slowly during the training. However, although this scheme avoids the conversion to FLP to calculate the exponent, some additional overhead is required to store the recorded statistics for each block. Thus, this scheme is suitable for the case when the number of blocks is relatively small. 

\begin{figure}[h]
    \centering
    \includegraphics[width=0.5\textwidth]{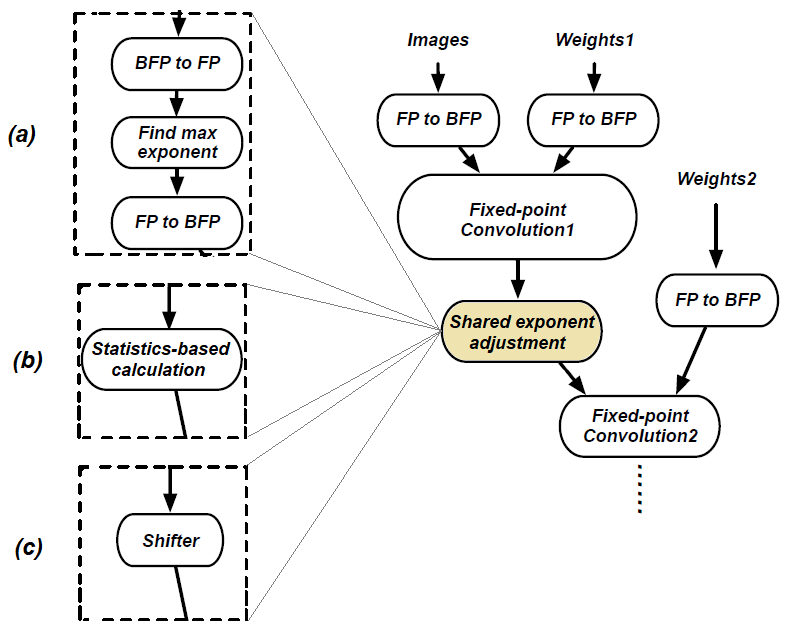}
    \caption{Shared exponent adjustment schemes: (a) Dynamic based on maximal exponent  (b) Dynamic based on statistics (c) Static. Adapted from \cite{fan2019static}}
    \label{shared}
\end{figure}

The static shared exponent scheme is presented to get rid of exponent calculation overhead when the BFP is employed for CNN inferences rather than training \cite{fan2019static, fan2021high, ni2020lbfp,darvish2020pushing}. Instead of dynamically calculating the shared exponent during run-time, the shared exponent can be set to a constant value estimated offline. The common approach to determine the shared exponent offline is to minimize the Kullback–Leibler (K-L) divergence \cite{claici2020model} between FLP32 distribution and BFP distribution of all blocks before the inference. By doing so, the extra memory and computational resources used for the exponent and the conversion between BFP and FLP are eliminated \cite{darvish2020pushing}. Because the input and output activations may have different shared exponents, a bit shifting is needed after each layer calculation, Figure \ref{shared}(c). Figure \ref{shared} summarizes the dataflow of the BFP when each of the three shared exponent determination schemes is adopted. 

\subsection{BFP Precision}
The precision of BPF is determined by the number of bits allocated for both the shared exponent and mantissa. Reducing this precision is an objective to increase the arithmetic efficiency and memory bandwidth. At the same time, the over-reduced bit-width of the mantissa results in what is known as zero setting problem 
\cite{noh2022flexblock}. This problem occurs when all the bits of the mantissa are shifted out resulting in a zero number representation, despite the presence of the exponent value. The over-reduction of the shared exponent number of bits is much worse. This is because of insufficiency to represent the actual exponent of the block, and thus the caused truncation ruins the correct representation of all numbers in the block.       

This precision is usually either static \cite{koster2017flexpoint,aydonat2017opencl,song2018computation,drumond2018training,fan2018reconfigurable,zhang2019block, lian2019high,yang2019swalp,fox2019training,fox2020block,fan2019static,su2020processing,ni2020lbfp,darvish2020pushing,fan2021high,wong2021low} or dynamic \cite{zhang2021fast,noh2022flexblock}. In the static precision, the number of bits is fixed and selected offline. To select the best precision, usually few experiments are performed using different number of bits \cite{yang2019swalp, su2020processing, fan2019static}. This gives an insight on the impact of this metric on the performance of DNN and allows for picking the minimum number of bits that preserve acceptable accuracy or the one that gives the best trade of between hardware efficiency and accuracy. Reducing the mantissa bit-width was paid attention in the literature because the performance of DNN is less sensitive to mantissa reduction compared to shared exponent. For example, 23-bit mantissa, same as the case of FLP, is required to guarantee the convergence of the Q-learning in \cite{su2020processing}, whereas 8-bit mantissa, or even less, was found to be sufficient for other CNN accelerator designs \cite{song2018computation,darvish2020pushing,zhang2019block,lian2019high, yang2019swalp, fox2019training, fox2020block, fan2019static, ni2020lbfp, darvish2020pushing, fan2021high, zhang2021fast, wong2021low, noh2022flexblock}. This indicates that the required static precision depends on the problem to be solved (mainly, the used dataset and DNN model).  

\begin{figure}[h]
    \centering
    \includegraphics[trim={0 0cm 0cm 0},clip,width=\linewidth]{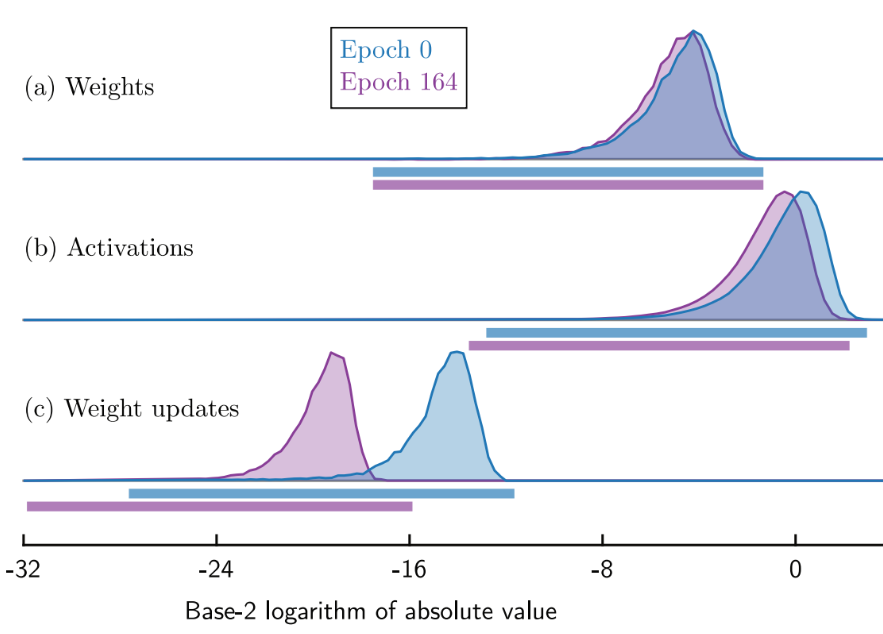}
    \caption{Example of how the distribution of weights (a), activations (b) and weight updates (c) changes during different DNN training iterations \cite{koster2017flexpoint}}
    \label{distribution}
\end{figure}

The dynamic precision of mantissas is presented in \cite{zhang2021fast,noh2022flexblock}. This dynamic precision is basically needed when the implemented DNN architecture is intended to be used for training rather than inference. This is attributed to the fact that the distributions of the weights, activations, and weight updates change during the training. Figure \ref{distribution} \cite{koster2017flexpoint} is an example of how the distribution changes during DNN training (at the start of the training and after 164 epochs). To speed up the training, the authors in \cite{zhang2021fast} proposed an adaptive training by changing the precision of BFP progressively across both training iterations and layer depth. This relies on the fact that the training is more amenable to low-precision in its early stages. In their approach, two levels of precision are supported, mainly 4-bit and 2-bit mantissas. For each block, the relative improvement due to using the higher precision is estimated by quantizing the block numbers using both precisions. Then, if this relative improvement is higher than a threshold the higher precision is used. This threshold differs based on the layer depth and training iteration. On the other hand, mixed dynamic precision of BFP is proposed because the distribution of the weight updates (gradients) changes more frequently than other variables during training \cite{noh2022flexblock}. This scheme assigns different, higher, precision to the weight updates compared to weights and activations. At the same time, their implementation supports adjusting this precision online during the training time to be one of the two levels (e.g., 4-bit or 8-bit mantissa). For each training iteration, the number of zero setting problem occurrences is tracked. If this problem happens more frequent than a predefined threshold, this indicates that the current precision is not sufficient and should be increased in the next training iteration. To avoid the fluctuation in the precision, a hysteresis controller is utilized by specifying two thresholds, upper and lower, for increasing and decreasing the precision, respectively. This dynamic precision showed no accuracy degradation with 16\% speed-up compared to the static precision. However, the dynamic precision advantage usually comes at the expense of added complication to the design of the hardware which should be reconfigurable with multi-mode arithmetic to adapt according to the selected precision.

\subsection{Summary and Discussion of BFP-based DNN architectures}

The main idea behind BFP representation is to strike a balance between the wide dynamic range but hardware inefficient FLP format and the limited-range hardware-friendly FXP format. BFP can be considered as a general format that has two extreme cases, i.e., the FLP case when each value is set in a separate block and the FXP case when the whole values of the architecture are treated as a single block with one shared exponent. Thus, different trade-offs can be obtained by specifying different design choices represented by the block size, shared exponent selection, and bit-width choice. Various CNN architectures that utilize BFP representation are listed in Table \ref{BFP_comp}. The first observation from this table is that even though BFP was initially proposed to implement efficient hardware capable of performing CNN training phase without ruining the accuracy, this representation got the same amount of attention for highly accurate inference hardware implementation. Most of these architectures achieved negligible accuracy degradation compared to FLP even with less than 8-bit mantissa \cite{fox2020block, darvish2020pushing,zhang2021fast}. Different implementations make use of different combinations of the discussed design choices, thus, the reported results of these works can't be used to prove the superiority of a specific design choice over the others. However, we can conclude that there is no clear trend in the accuracy enhancement when tile-based blocking is used instead of a filter-based one.    

\begin{table*}[!htbp]\fontsize{6pt}{12}\selectfont
	\caption{Comparison of DNN architectures based on BFP representation}
	\centering
	\label{BFP_comp}
\begin{tabular}{|c|c|c|c|c|c|c|c|c|c|c|c|}
\hline
Ref.    & Phase                                                             & DNN                                                                & \begin{tabular}[c]{@{}c@{}}Exponent\\ selection\end{tabular}  & \begin{tabular}[c]{@{}c@{}}Block\\ design\end{tabular} & Dataset                                                                   & Model                                                    & \begin{tabular}[c]{@{}c@{}} Mantissa\\ bits\end{tabular} & \begin{tabular}[c]{@{}c@{}}Accuracy\\ loss \%\end{tabular} & \begin{tabular}[c]{@{}c@{}}Area\\ saving \%\end{tabular}           & \begin{tabular}[c]{@{}c@{}}Power\\ saving \%\end{tabular}                     & \begin{tabular}[c]{@{}c@{}}Speed\\ up (times)\end{tabular}                          \\ \hline
\cite{koster2017flexpoint}  & \begin{tabular}[c]{@{}c@{}}Training,\\ inference\end{tabular}     & \begin{tabular}[c]{@{}c@{}}CNN,\\ GANs\end{tabular}                & \begin{tabular}[c]{@{}c@{}}Dynamic,\\ statistics\end{tabular} & Filter-based                                           & \begin{tabular}[c]{@{}c@{}}CIFAR-10,\\  LSUN\end{tabular}                 & \begin{tabular}[c]{@{}c@{}}AlexNet,\\ WGAN\end{tabular}                                              & 16                                                          & $\sim$0                                                 & -                                                               & -                                                                          & -                                                                           \\ \hline
\cite{aydonat2017opencl} & Inference                                                         & CNN                                                                & \begin{tabular}[c]{@{}c@{}}Dynamic,\\ max\end{tabular}    & Tile-based                                             & ILSVRC                                                                    & AlexNet                                                                                              & 10                                                          & $\sim$0                                                 & -                                                               & -                                                                          & \begin{tabular}[c]{@{}c@{}}10\\ (\cite{zhang2018caffeine})\end{tabular}           \\ \hline
\cite{song2018computation}    & Inference                                                         & CNN                                                                & \begin{tabular}[c]{@{}c@{}}Dynamic,\\ max\end{tabular}    & Filter-based                                           & \begin{tabular}[c]{@{}c@{}}MNIST,\\ CIFAR10\end{tabular}                  & \begin{tabular}[c]{@{}c@{}}VGG16,\\ ResNet-18,\\ ResNet-50,\\ GoogLeNet\end{tabular}                 & 8                                                           & \textless{}0.3                                        & -                                                               & -                                                                          & -                                                                           \\ \hline
\cite{drumond2018training} & \begin{tabular}[c]{@{}c@{}}Training,\\ inference\end{tabular}     & \begin{tabular}[c]{@{}c@{}}CNN,\\ RNN\end{tabular}                 & \begin{tabular}[c]{@{}c@{}}Dynamic,\\ max\end{tabular}    & Tile-based                                             & \begin{tabular}[c]{@{}c@{}}CIFAR-100,\\ SVHN,\\ ImageNet\end{tabular}     & \begin{tabular}[c]{@{}c@{}}ResNet,\\ WideResNet,\\ DenseNet\end{tabular}                             & 8                                                           & \textless{}1                                          & -                                                               & -                                                                          & \begin{tabular}[c]{@{}c@{}}8.5\\ (FLP16)\end{tabular}                       \\ \hline
\cite{fan2018reconfigurable}     & Inference                                                         & CNN                                                                & \begin{tabular}[c]{@{}c@{}}Dynamic,\\ max\end{tabular}    & Tile-based                                             & Sports-1M                                                                 & Custom                                                                                               & 15                                                          & 0.4                                                   & -                                                               & \begin{tabular}[c]{@{}c@{}}92\\ (Intel i7-950)\end{tabular}              & \begin{tabular}[c]{@{}c@{}}8.2\\ (Intel i7-950)\end{tabular}               \\ \hline
\cite{zhang2019block}   & Inference                                                         & CNN                                                                & \begin{tabular}[c]{@{}c@{}}Dynamic,\\ max\end{tabular}    & Filter-based                                           & ImageNet                                                                  & \begin{tabular}[c]{@{}c@{}}VGG-16,\\ GoogLeNet,\\ ResNet-50\end{tabular}                             & 8                                                           & \textless{}0.14                                       & -                                                               & \begin{tabular}[c]{@{}c@{}}31 \\ (FLP16 {\cite{mei2017200mhz}})\end{tabular} & -                                                                           \\ \hline
\cite{lian2019high}    & Inference                                                         & CNN                                                                & \begin{tabular}[c]{@{}c@{}}Dynamic,\\ max\end{tabular}    & Filter-based                                           & \begin{tabular}[c]{@{}c@{}}ImageNet,\\ CIFAR10\end{tabular}               & \begin{tabular}[c]{@{}c@{}}LeNet,\\ VGG-16,\\ GoogLeNet,\\ ResNet-50\end{tabular}                    & 8                                                           & 0.12                                                 & -                                                               & \begin{tabular}[c]{@{}c@{}}15\\  (FLP16 \cite{mei2017200mhz})\end{tabular} & \begin{tabular}[c]{@{}c@{}}3.76 \\ (FLP16 \cite{mei2017200mhz})\end{tabular} \\ \hline
\cite{yang2019swalp}    & \begin{tabular}[c]{@{}c@{}}Training,\\ inference\end{tabular}     & CNN                                                                & \begin{tabular}[c]{@{}c@{}}Dynamic,\\ max\end{tabular}    & Filter-based                                           & \begin{tabular}[c]{@{}c@{}}ImageNet,\\ CIFAR100,\\ CIFAR10\end{tabular}   & \begin{tabular}[c]{@{}c@{}}VGG16,\\ PreResNet-164,\\ logistic reg.\end{tabular}                & 8                                                           & \textless{}3                                          & -                                                               & -                                                                          & -                                                                           \\ \hline
\cite{fox2019training}     & \begin{tabular}[c]{@{}c@{}}Training,\\ inference\end{tabular}     & CNN                                                                & \begin{tabular}[c]{@{}c@{}}Dynamic,\\ max\end{tabular}    & Filter-based                                           & \begin{tabular}[c]{@{}c@{}}MNIST,\\ CIFAR10\end{tabular}                  & \begin{tabular}[c]{@{}c@{}}VGG16,\\ PreResNet-164\end{tabular}                                       & 8                                                           & 0.1                                                   & -                                                               & -                                                                          & \begin{tabular}[c]{@{}c@{}}17\\ (ARM A53 )\end{tabular}                    \\ \hline
\cite{fox2019training}     & \begin{tabular}[c]{@{}c@{}}Training,\\ inference\end{tabular}     & \begin{tabular}[c]{@{}c@{}}CNN,\\ RNN\end{tabular}                 & \begin{tabular}[c]{@{}c@{}}Dynamic,\\ max\end{tabular}    & Filter-based                                           & \begin{tabular}[c]{@{}c@{}}ImageNet,\\ CIFAR-10,\\ CIFAR-100\end{tabular} & \begin{tabular}[c]{@{}c@{}}ResNet,\\ LENET\end{tabular}                                              & \textless{}=8                                               & \textless{}1                                          & \begin{tabular}[c]{@{}c@{}}92-96\end{tabular}    & \begin{tabular}[c]{@{}c@{}}91\end{tabular}     & -                                                                           \\ \hline
\cite{fan2019static}     & Inference                                                         & CNN                                                                & Static                                                        & Filter-based                                           & ImageNet                                                                  & \begin{tabular}[c]{@{}c@{}}ResNet–50,\\ VGG–16,\\ Inception,\\ MobileNet\end{tabular}                & 8                                                           & \textless{}1                                         & \begin{tabular}[c]{@{}c@{}}$\sim$50\\ (FLP8 MAC)\end{tabular}  & -                                                                          & -                                                                           \\ \hline
\cite{su2020processing}      & \begin{tabular}[c]{@{}c@{}}Training,\\ inference\end{tabular}     & DRL                                                                & \begin{tabular}[c]{@{}c@{}}Dynamic,\\ statistics\end{tabular} & Filter-based                                           & -                                                                         & Q-learning                                                                                           & 23                                                          & -                                                       & \begin{tabular}[c]{@{}c@{}}15.8\end{tabular}     & -                                                                          & -                                                                           \\ \hline
\cite{ni2020lbfp}      & Inference                                                         & CNN                                                                & Static                                                        & Filter-based                                           & ImageNet                                                                  & \begin{tabular}[c]{@{}c@{}}ResNet-18,\\ ResNet-50\end{tabular}                                       & 8                                                           & \textless{}0.6                                       & \begin{tabular}[c]{@{}c@{}}15\\ (FXP8 MAC)\end{tabular}       & \begin{tabular}[c]{@{}c@{}}16\\ (FXP8 MAC)\end{tabular}                  & -                                                                           \\ \hline
\cite{darvish2020pushing} & \begin{tabular}[c]{@{}c@{}}Inference\\ (fine tuning)\end{tabular} & \begin{tabular}[c]{@{}c@{}}CNN,\\ RNN\end{tabular} & Static                                                        & Tile-based                                             & ImageNet                                                                  & many                                                                                                 & \textless{}=8                                               & \textless{}1                                          & \begin{tabular}[c]{@{}c@{}}88-97\end{tabular}    & -                                                                          & -                                                                           \\ \hline
\cite{fan2021high}     & Inference                                                         & CNN                                                                & Static                                                        & Filter-based                                           & ImageNet                                                                  & \begin{tabular}[c]{@{}c@{}}ResNet–50,\\ VGG–16,\\ Inception, \\ MobileNet\end{tabular}               & 8                                                           & \textless{}1                                          & \begin{tabular}[c]{@{}c@{}}$\sim$50\\ (FXP8 MAC)\end{tabular} & \begin{tabular}[c]{@{}c@{}}82\\ (TITAN GPU)\end{tabular}                 & \begin{tabular}[c]{@{}c@{}}1.13\\  (TITAN GPU)\end{tabular}                \\ \hline
\cite{zhang2021fast}   & \begin{tabular}[c]{@{}c@{}}Training,\\ inference\end{tabular}     & CNN                                                                & \begin{tabular}[c]{@{}c@{}}Dynamic,\\ max\end{tabular}    & Tile-based                                             & ImageNet                                                                  & Custom                                                                                               & 3                                                           & \textless{}2                                          & -                                                               & \begin{tabular}[c]{@{}c@{}}80\\ (FLP16)\end{tabular}                      & \begin{tabular}[c]{@{}c@{}}2\\ (\cite{darvish2020pushing})\end{tabular}                   \\ \hline
\cite{wong2021low}    & Inference                                                         & CNN                                                                & \begin{tabular}[c]{@{}c@{}}Dynamic,\\ max\end{tabular}    & Tile-based                                           & \begin{tabular}[c]{@{}c@{}}Torchvision,\\ ImageNet\end{tabular}           & \begin{tabular}[c]{@{}c@{}}VGG16,\\ ResNet-152\end{tabular}                                          & 10                                                          & \textless{}1                                         & \begin{tabular}[c]{@{}c@{}}50.1\\ (FLP16 LUTs)\end{tabular}    & -                                                                          & \begin{tabular}[c]{@{}c@{}}1.32\\ (FLP16)\end{tabular}                      \\ \hline

\end{tabular}

\end{table*}

\section{DFXP for DNN Architectures}\label{DFXP_sec}
DFXP representation shares the same concept of BFP discussed in Section \ref{BFP} and sometimes the notations DFXP and BFP are used interchangeably. As in the case of BFP, in DFXP, the values are grouped and different scaling factors (i.e., shared exponents) are used for different groups. Thus, a scaling factor is unique for each group (e.g., layer). In some cases, it can be changed from time to time (i.e., dynamic). This is compared to the case of FXP which assigns a single global scaling factor for the whole DNN architecture all the time. To this end, Equations (\ref{tensor},\ref{tensor_B},\ref{tensor_dot}) are applicable to DFXP. Although several works use the term DFXP to indicate a representation similar to BFP \cite{sakai2020quantizaiton, jo2018training, das2018mixed, wu2019efficient}, the majority of works use DFXP to indicate FXP representation provided with flexibility to change the place of the decimal point, that specifies the length of the integer and fraction parts for each group of values, Figure \ref{dfxp_rep}. This requires that a scaling factor $\epsilon_{N}$ of a group $N$, (\ref{tensor_B}), to be in the range $[-w_N,0]$, where $w_N$ is the bit-width used to represent elements of a group $N$ \cite{Prado2018quenn,lin2019design, guo2020hybrid}. Hence, DFXP representation can be reduced to $<I_N,F_N>$ format where $I_N,\ F_N$ are the number of bits allocated to the integer and fractional parts, respectively, for all values within a group. Such that $w_N=I_N+ F_N$ and $\epsilon_{N}=-F_N$. Thus, the zero setting problem frequently happens with BFP will not appear for the DFXP at the expense of limited dynamic range, but still better than the one of FXP. We will limit our discussion on these works in this section, whereas the other works that use DFXP notation to indicate BFP are discussed in Section \ref{BFP} although they use the term DFXP.

A notable difference between DNN architectures that use BFP and DFXP is that the latter gives less attention to the way that the groups (i.e., blocks) are partitioned. The common grouping approach for DNN architecture based on DFXP is to consider the weights, biases, input activation, and gradients vectors (when DFXP is used to accelerate training) for each layer as separate groups and thus associated with different scaling factors \cite{courbariaux2014training, gysel2016hardware,na2016speeding,shan2016dynamic,peng2017running, jo2018training, lai2017deep, Prado2018quenn, shin2017energy,han2018low}. Only one architecture presented in \cite{mellempudi2017mixed} statically clusters the filters (i.e., weights) that accumulate to the same output activation of each layer. Then, each cluster represents a group that has its unique scaling factor. The quantization error is effectively reduced with smaller clusters (e.g., when a cluster contains 4 filters) since smaller groups tend to have smaller range of values.

The main differences between the DFXP representation in different works are the way of finding the best scaling factor $F_N$ and determining the bit-width $w_N$. The approaches used to optimize the decimal point Position and specify the precision of DFXP are classified in the next subsections.

\subsection{Group Scaling Factor Selection}

The scaling factor (i.e., $F_N$) assignment to each group in DFXP is usually performed in an offline or online manner. The offline assignment is usually used when the architecture is implemented for inference purpose \cite{gysel2016hardware,na2016speeding, shan2016dynamic,lai2017deep,shin2017energy,mellempudi2017mixed,peng2017running, Prado2018quenn, wu2019efficient,shawahna2022fxp, lin2021hybrid,kuramochi2020fpga, liu2019implementation, ding2019fpga, mitschke2019fixed, lin2019design }. The common approach for the  offline assignment depends on finding the minimum integer bit-width $I_N$ that accommodates the maximum value within a group as in 
 \begin{equation} \label{I_N}
I_N= \log_2(\max(|n_{max}|,|n_{min}|)),  
\end{equation}

where $n_{max}\text{, }n_{min}$ are the maximum and minimum values within a group $N$. The remaining bits $w_N-I_N$ are allocated to the fractional part $F_N$. This approach is used for example in \cite{gysel2016hardware,lin2019design}. However, as the presence of outliers in a group results in an unnecessary increase in the integer bit-width, the outliers can be excluded before calculating the bit-width $I_N$ \cite{lin2019design}. Several works minimize the impact of the outliers by selecting a scaling factor that minimizes the error between computed and real values \cite{guo2020hybrid,shawahna2022fxp,shan2016dynamic}. For instance, K-L divergence between FLP32 and DFXP weight distributions is used in \cite{guo2020hybrid}, whereas a greedy algorithm is utilized in \cite{shan2016dynamic} to determine the best scaling factor.

\begin{figure}[h]
    \centering
    \includegraphics[width=0.5\textwidth]{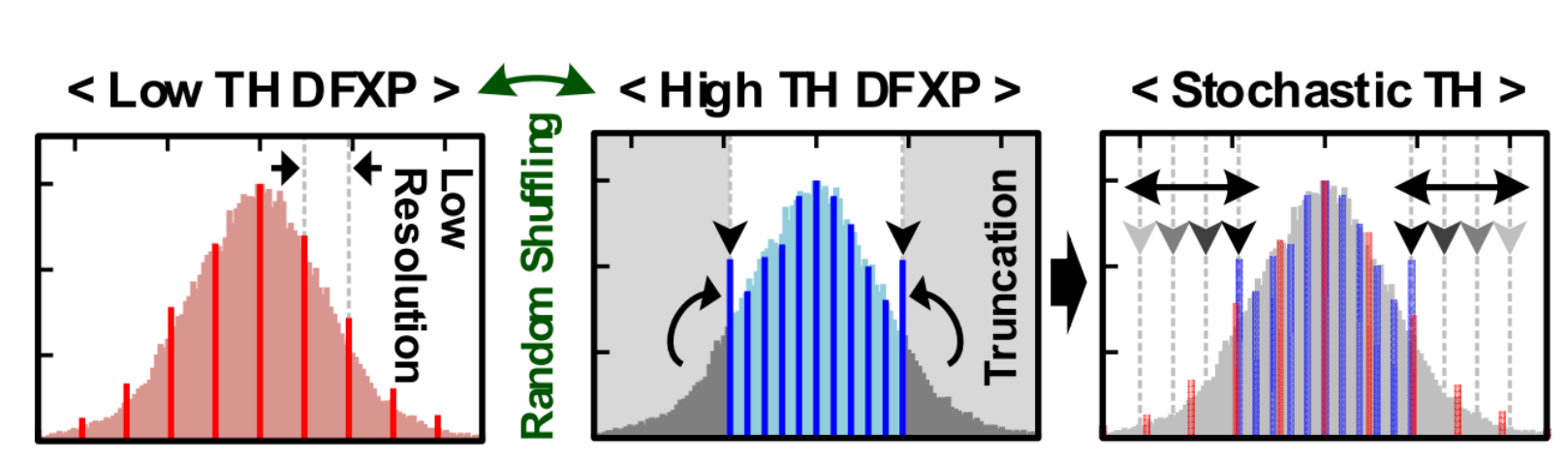}
    \caption{Comparing the deterministic and stochastic thresholds for online scaling factor selection \cite{han2021hnpu}}
    \label{stochastic_thresholding}
\end{figure}

The online scaling factor selection is needed for the training phase in which the values within each group change frequently \cite{courbariaux2014training, na2016speeding, na2017chip, shin2017energy, taras2018quantization, han2021hnpu, guo2020hybrid, sakai2020quantizaiton, taras2018quantization}. Usually, the scaling factor is updated at a given frequency based on the rate of overflow during the training. When the current integer part fails to handle a value in a group, the overflow rate increases. The overflow rate is compared to a threshold to decide whether this scaling factor should be increased or decreased. This threshold can be deterministic and predefined \cite{courbariaux2014training,na2016speeding,taras2018quantization,na2017chip,shin2017energy}, or stochastic \cite{han2021hnpu}. The stochastic thresholding is presented because the lower deterministic threshold results in inaccurate representation for small values while the higher threshold causes large clipping error \cite{han2021hnpu}, Figure \ref{stochastic_thresholding}. The random shuffling between higher and lower thresholds is found to be effective in compensating for the accuracy degradation of the low-precision training (less than 6 bits).      

\subsection{DFXP Precision}
The bit-precision of DFXP (i.e., $w_N$) can be static, mixed, or dynamic with different trade-offs between accuracy and hardware efficiency. The static precision, which is used in \cite{na2017chip,mellempudi2017mixed, ding2019fpga, han2018low, lin2019design,lin2021hybrid, guo2020hybrid}, indicates that the number of bits is statically specified prior and is kept fixed for all groups during the training or inference, i.e., $w_{N_i}=w_c \text { for } i= 1, \dots, N_t$, where $N_t$ is the total number of groups associated with a specific DNN architecture. The advantage of this scheme is its simplicity from the hardware efficiency point of view. However, the selected precision is not optimal for all groups, layers, and architectures \cite{han2021hnpu}. 

On the other hand, in the mixed-precision scheme, the bit-width, which is determined offline as well, can be different for different groups \cite{mellempudi2017mixed, Prado2018quenn}. The need for mixed-precision mainly comes from the fact that different groups (such as weights and activations) have different required dynamic ranges and thus different required number of bits \cite{liu2019implementation}. As the activation results from the convolution accumulation, it is usually allocated more bits. For instance, using DFXP with 4-bit wights and 8-bit activations gives an accuracy degradation within 2\% of the full precision using the Resnet-50 CNN model on the ImageNet dataset \cite{mellempudi2017mixed}. In other works, different precision is allocated to different groups in different layers \cite{Prado2018quenn,shawahna2022fxp}. The authors in \cite{Prado2018quenn} stated that a specific fully connected layer activation is more sensitive to bit reduction and it is better to be allocated 16 bits while the activation bit-width of the other layers can be shrink to 8 bits. This mixed-precision allows them to achieve a 55.64\% saving for weights' storage and 69.17\% for activations’ memory traffic with less than 2.5\% loss in the accuracy when the Alexnet model and ImageNet dataset are used. The experiments in \cite{shawahna2022fxp} show similar results. They found that the groups in shallower layers are less robust to bit reduction than the ones in deeper layers. In addition, the computation of the first and the last network layers should use high bit-precision to achieve better performance. To optimize the mixed-precision for different groups and to reach the above conclusions, the authors in \cite{shawahna2022fxp} adopted an iterative bit-precision reduction scheme that aims to discover the groups for which the bit precision can be reduced without causing noticeable performance degradation. When DFXP with mixed-precision is used for training, sometimes different bit-widths are used for the weights during the updates than during the forward and backward propagations \cite{courbariaux2014training}. Using higher precision for the weight updates allows for the small changes in the weights to be accumulated precisely. 

The use of DFXP with dynamic precision is presented to adjust the bit-width on-the-fly during training to enable speeding up this process \cite{na2016speeding, taras2018quantization, han2021hnpu}. The scheme in \cite{na2016speeding, taras2018quantization} suggests starting with an aggressive initial target bit-width and monitoring the training loss as a feedback from the training process. If the training becomes unstable, the bit-width is increased to its maximum value. Afterward, the target bit-width is gradually increased by a unit step for the next trial. This procedure is repeated until reaching the minimum target bit-width that allows for stable training. To maintain the low overhead of this algorithm, it is activated once after each forward/backward computation to find the global bit-width of DNN architecture. A simpler search-based scheme to adapt the bit-width of each layer is suggested in \cite{han2021hnpu}. In this scheme, the convolution is calculated in presence of low and high bit-widths at the same time for several iterations per epoch. If the difference between the high and low precisions is higher than a predefined threshold, the bit-width increases starting from the next iteration till the end of the epoch. After applying this scheme to different datasets and different CNN models, an interesting conclusion was that different datasets require different average bit-widths even if the same model is used. 

One added complication of utilizing the dynamic bit-width is the need to design a configurable processing unit that can be configured to compute with various bit-widths during run-time. Thus, the efficiency of the dynamic precision scheme is highly affected by the hardware’s supportability of the bit-width levels. Two relatively high bit-width levels (32 bits and 64 bits) are adopted in \cite{na2016speeding}. The baseline precision to prove the efficiency of their proposed approach is 64 bits which is relatively high training precision compared to other works. On the other hand, \cite{han2021hnpu} could train the CNN with negligible loss of training and testing accuracy using an average bit-width less than 8. This is because they were able to use finer bit-width levels thanks to the bit-slice serial architecture they proposed.

\subsection{Summary and Discussion of DFXP-based DNN Architectures}

DFXP and BFP are very similar representations. DFXP can be considered as a subset of BFP with less dynamic range and less hardware complication at the same time. For example, when the DFXP representation is used for CNN inference, the only additional hardware required over the FXP is a simple bit-shifter to align the output activation with the scale factor of the next layer input activation \cite{lin2021hybrid, liu2019implementation}. This simplicity makes it appealing for many DNN architectures \cite{gysel2016hardware, shan2016dynamic, mellempudi2017mixed, peng2017running,courbariaux2014training, na2017chip, jo2018training, wu2019efficient, shin2017energy, na2016speeding, das2018mixed, Prado2018quenn, lo2018fixed, lai2017deep, han2021hnpu, shawahna2022fxp, sakai2020quantizaiton, mitschke2019fixed, lin2019design, guo2020hybrid, lin2021hybrid, liu2019implementation, han2018low, ding2019fpga, kuramochi2020fpga, yang2021fixar, taras2018quantization, kang2018weight}. By considering the number of accelerators in the literature that utilize each representation, DFXP can be considered the most widely used alternative number system. The widespread of DFXP can be attributed to its simplicity and to implementing it in some of publicly available DNN frameworks, such as Ristretto \cite{gysel2016hardware}. Several of the DFXP-based DNN architectures used this representation without much contribution to the proposed vanilla DFXP. Other works used different approaches to select the scaling factor of each group and to optimize the bit-width of this representation. Different approaches used for these design metrics are discussed and compared above. 

\section{Posit for DNN Architectures}\label{Posit_sec}
Posit number system, also known as type III universal number (Unum) system \cite{lu2020evaluations,carmichael2019deep,gustafson2017beating}, is a floating-like format that is proposed to overcome several shortcomings of FLP representation \cite{carmichael2019deep}. Compared to FLP, Posit uses the bits more efficiently (allows for better accuracy with the same number of bits) \cite{cococcioni2019fast}, and has better accuracy and dynamic range \cite{carmichael2019deep, romanov2021analysis}. 
Figure \ref{Posit_rep} illustrates Posit representation. The $w$ bits Posit number representation consists of four fields; a sign (1 bit), a regime (of variable length $rs \in [1,w-1]$), an exponent $e$ (unsigned integer of fixed length $es$) and a mantissa (of variable length $ms=w-rs-es-1$). The regime field contains $d$ consecutive identical bits and an inverted terminating bit (i.e., $rrr \dots \bar{r}$)\footnote{This is the general case when $rs<(w-1)$. Otherwise, the regime pattern can be $rrr \dots {r}$ when it is terminated by the end of the $w$ bits \cite{langroudi2020adaptive}.
}. 

The numerical value of a real number $n$ is represented in the Posit format (by $\hat{n}$ ) as follows:
 \begin{equation} \label{Posit_format}
\hat{n} =(-1)^s \times u^k \times 2^e \times (1+\frac{m}{2^{ms}}),
\end{equation}
where $s\text{, } e\text{, and } m  $ are the values of the sign bit, exponent and mantissa, respectively. The useed $u\text{, and } k$  are calculated in (\ref{useed}) and (\ref{kkk}) with the same order. 
 \begin{equation} \label{useed}
u= 2^{2^{es}},  
\end{equation}
\begin{align}\label{kkk}
k =
\left\{
\begin{array}{ll}
-d, &  \text{if }  r=0.\\
d-1, & \text{if } r=1,
\end{array}
\right.
\end{align}

Posit representation is commonly characterized by two parameters, mainly $w$ and $es$, and defined as Posit($w$,$es$) \cite{lu2020evaluations,langroudi2018deep}. The parameter $es$ is used to control the trade-off between the precision and the dynamic range \cite{lu2020evaluations}. When the Posit is intended to be used for DNN, these parameters are usually specified in an offline manner regardless of whether this architecture targets DNN training or inference \cite{lu2020evaluations, langroudi2018deep, montero2019template, carmichael2019deep}. The selection of these parameters is done usually by experimenting with different parameters and selecting the parameters that give the best accuracy \cite{montero2019template} or the parameters that offer the best balance between the accuracy and hardware efficiency. For instance, when the exponent length is set to $es=1$ in \cite{carmichael2019performance} a better trade-off between accuracy and energy-delay-product is obtained for $w=7 \text{ and } w=5$. On the other hand, the author in \cite{langroudi2018deep} decided to eliminate the exponent part (i.e., $es=0$) as the Posit, in this case, better represents the dynamic range of the used DNN weights.

\begin{figure}[h]
    \centering
    \includegraphics[width=0.5\textwidth]{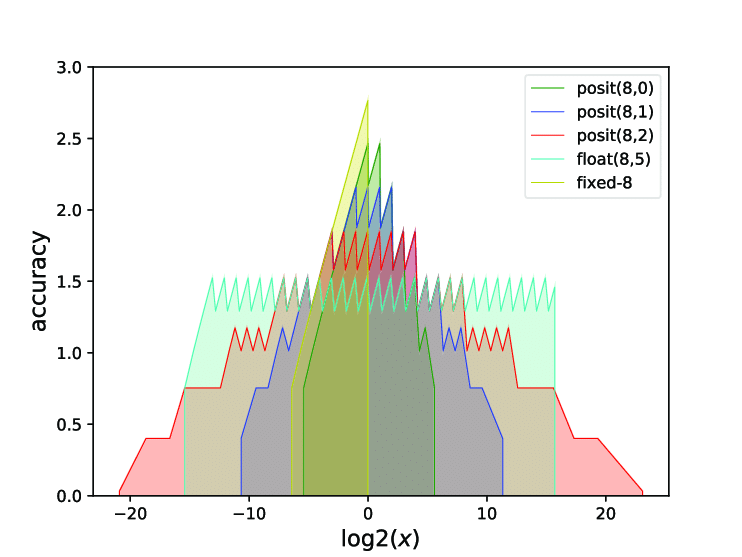}
    \caption{The decimal accuracy of the Posit representation compared to the FLP8 and FXP8 \cite{lu2020evaluations}}
    \label{Posit_typ}
\end{figure}

There are two main differences between Posit and FLP representations, Figure \ref{Posit_rep} and \ref{fp_rep}. The first difference is the presence of the regime field, and the second is the variability of the mantissa bit-length. Indeed, the innovation in the Posit format comes from its ability to allocate more bits to the mantissa when the represented number is very small (i.e., higher precision) and fewer bits for large numbers (i.e., larger magnitude) without changing the total bit-width of the format \cite{cococcioni2019fast}. The Posit is usually known for its \textit{tapered-accuracy}, i.e., small magnitude numbers around the '1' have more accuracy than extremely large or extremely small numbers \cite{langroudi2021alps}. The authors in \cite{lu2020evaluations} compared the decimal accuracy ( $-\log_{10}|\log_{10} (\frac{\hat{x}}{x})|$, where $x$ is the actual real number value and $\hat{x}$ is the represented number value \cite{gustafson2017beating}) of different Posit representations to the FLP8 and FXP8, see Figure \ref{Posit_typ}. Their experiment showed that: i)  the FXP representation has a peak accuracy so it is suitable to represent data with a narrow range, ii) the floating point has almost constant accuracy and it should be used to represent data that are uniformly distributed to exploit its efficiency, iii) and the Posit has tapered accuracy which makes it suitable to represent the normally distributed data efficiently. Since data in DNNs usually are normally distributed, see for example Figure \ref{distribution}, Posit is expected to be the most attractive number system for DNN \cite{lu2020evaluations}. 

 DNN architectures that use Posit number system are usually either rely on Posit format from end-to-end \cite{wan2018study, romanov2021analysis, murillo2020customizedMAC,cococcioni2019fast,murillo2021plam} or partially utilize this format and a conversion from and to other formats is required within this architecture \cite{nambi2021expan, langroudi2018deep}. These two approaches of using Posit are discussed next. In addition, to increase the efficiency of Posit number system for DNNs, several Posit variants are proposed. These variants are reported below as well.

\subsection{End-to-end Posit-based Architectures}
When DNN data are represented in Posit from end-to-end new hardware that is able to perform all operations on these data must be used. In this case, the most fundamental arithmetic operations that need to be carefully designed in hardware are MAC operation and activation functions \cite{cococcioni2019fast}. 

Different designs of the Posit-based MAC (or multiplier) are proposed in \cite{murillo2021plam, gohil2021fixed, murillo2020deep, carmichael2019deep, carmichael2019performance,zhang2019efficient, langroudi2019Positnn,podobas2018hardware,lu2020evaluations}. In most of these works, the MAC design mainly follows the standard FLP MAC as in \cite{murillo2020deep, carmichael2019deep, carmichael2019performance,zhang2019efficient, langroudi2019Positnn,podobas2018hardware,lu2020evaluations}. The main additional steps over the FLP MAC design are the decoding to extract Posit fields of the operands and encoding the result to Posit format \cite{murillo2021plam}. Indeed, Posit MAC hardware implementation is more complicated and less efficient than the FLP MAC with the same number of bits because of the length-variability of the regime and mantissa fields. It is shown in \cite{gohil2021fixed} that Posit(32, 6) multiplier has 78\% more area and consumes 94\% more power than the FLP32 multiplier. This is attributed to the fact that the multiplier should be designed to handle the extreme lengths of mantissa, which is $w-es-2$, and regime, which is $w-1$. In addition, the critical path of this Posit multiplier is found to be longer than FLP32 due to the sequential bit decoding required for Posit. By making the fields of Posit format fixed, the area and power efficiency increased by 47\%, and 38.5\%, respectively, over the variable length fields Posit at the expense of negligible accuracy loss. Similar results are shown in \cite{walia2021fast} as well.  

Alternatively, to design a more power and area-efficient multiplier, the authors in \cite{murillo2021plam} proposed a Posit-LNS-Approximate multiplication. This combination allows for exploiting the advantages of Posit accuracy and LNS hardware efficiency. The general concept of performing LNS multiplications is similar to Mitchell's approximation discussed in section \ref{Mitchell_sec}, however, by considering Posit format instead. For example, the logarithm of a Posit number is given in (\ref{Posit_lns}) by taking the logarithm of both sides of (\ref{Posit_format}) and applying the approximation in (\ref{log_approxi}). 

 \begin{equation} \label{Posit_lns}
\log_2(|\hat{n}|) = 2^{es}\times k+e+  \frac{m}{2^{ms}}.
\end{equation}
Consequently, Posit multiplication is performed using fixed point addition. The experiments in \cite{murillo2021plam} showed significant reduction in the multiplier area by 72.86\%, power by 81.79\%, and delay by 17.01\% compared to Posit multipliers in \cite{murillo2020customizedMAC}.  

The implementation of several activation functions of Posit represented data is discussed in \cite{cococcioni2020fast, lu2020evaluations,wan2018study}. The Sigmoid activation function in (\ref{sig}) is found to be easy to be implemented in hardware for Posit represented data \cite{gustafson2017beating}. Few simple bit-cloning and masking is adequate to approximate this function. Similarly, a fast implementation of the  $\tanh$ and the Extended Linear Unit (ELU) activation functions are presented in \cite{cococcioni2020fast} and \cite{cococcioni2020novel}, respectively. 

\subsection{Partial Posit-based Architectures}

Several architectures aimed to benefit from the high accuracy and dynamic range of Posit while avoiding its hardware inefficiency by representing only the weights with Posit prior to the inference process \cite{nambi2021expan, langroudi2018deep}. This enables significant decrease in both the storage and communication overheads. These weights are then converted back to another format, such as FLP in \cite{langroudi2018deep} or FXP in \cite{nambi2021expan}, during the computation. The only overhead over the hardware of the standard architectures are modules to convert from Posit to the other formats and vice versa. The penalty of converting Posit to FXP is the increase in critical path delay and power consumption of the MAC by 22.8\% and 5\%, respectively \cite{nambi2021expan}.

\subsection{Posit Variants}
Two Posit variants are proposed for DNNs; the fixed-Posit representation \cite{gohil2021fixed}, and the generalized Posit representation \cite{langroudi2020adaptive, langroudi2021alps}. As its name indicates, the fixed-Posit representation proposes using a fixed length of the regime $rs=constant$ instead of using a variable length in the vanilla Posit. Although the dynamic range and the accuracy of this representation are expected to be less than that of Posit, using this representation results in much more efficient hardware, in terms of power, area, and delay, with negligible loss in classification accuracy (0.12 \%) when it used for ResNet-18 on ImageNet \cite{gohil2021fixed}.

The generalized Posit representation \cite{langroudi2020adaptive, langroudi2021alps} proposed a modification to the vanilla Posit format to better represent the dynamic range and data distribution of DNNs. They relied on the fact that Posit with $w<8$ and a specific $es$ is observed to be unable to accommodate the variability in parameter distributions and dynamic range of different DNN layers and various DNN models. Instead of using mixed-precision Posits which requires a very large search space (as huge as $4^{110}$ for ResNet-110 when 4 different $w$ values are searched \cite{langroudi2020adaptive}), Posit format is modified by inserting two hyper-parameters that can be adjusted per-layer to enable a parameterized tapered accuracy and dynamic range. These two hyper-parameters are the exponent bias and the maximum regime bit-width that can be applied by replacing $e $ in (\ref{Posit_format}) with $e+sc$, where $sc$ is the exponent bias, and restricting the number of bits allocated to the regime $rs\leq rs_{max}$. The exponent bias is used to scale the zone of maximum accuracy (i.e., minimum and maximum magnitude values) downward or upward in order to track the data distribution of different layers. The maximum regime bit-width $rs_{max}$ controls the maximum and minimum Positive representable values. When $rs_{max}=1$, the generalized Posits becomes a FLP-like format, whereas it turns into vanilla Posit format with $rs_{max}=w-1$. Various tapered-precision representations can be obtained by selecting the $rs_{max}$ between these two bounds. The experimental results on several datasets and CNN models showed that the generalized Posit offers considerable accuracy improvement when $w<8$ bits compared to the vanilla Posit at the expense of a relatively moderate increase in energy consumption.

\subsection{Summary and Discussion of Posit-based DNN Architectures}

Posit representation can be considered as a variant of FLP. This representation offers better accuracy and wider dynamic range than FLP. Thus, Posit can represent DNN data more efficiently with the same number of bits. However, in general, the hardware implementation of Posit is found to be more complicated compared to FLP hardware, as it relies on the FLP hardware in addition to the hardware needed to convert from and to FLP. Several trials have been made to enhance Posit hardware efficiency discussed above such as combining Posit with other representations (FXP and LNS) or modifying Posit by fixing or limiting the regime field.


\section{Future Directions and Open Research Issues}\label{Future}
Next, we briefly highlight several issues and opportunities for future research in DNN number systems. This includes dynamic number representations, hybrid number systems, and utilization of DNN statistics .

\subsection{Dynamic Number Systems} 
The main challenge of using low-precision number systems for training DNNs is the dynamic distribution of weights, activation, and gradients during training. In addition, several works show that optimal parameters of the number system (e.g., bit-widths) can be different for different datasets. This makes a dynamic number system (i.e., a number system that can adjust its parameters either offline or during run-time) highly desirable, especially for training DNNs. However, implementing such a system with online adaptation adds complications to the hardware which should be re-configurable to adapt to the changes in the number system format. Several works that adopt a format with a dynamic bit-width, for example \cite{zhang2020fixed}, discussed the worthiness of this approach from a software (accuracy and speed gain) point of view. it seems worthy to investigate the effectiveness of a dynamic number system from the hardware efficiency perspective.  

\subsection{Hybrid Number Systems}
Several hybrid number systems have been investigated. Some examples of hybrid representations include DFXP with binary FXP \cite{guo2020hybrid}, DFXP with ternary FXP \cite{mellempudi2017mixed}, DFXP with FLP \cite{lai2017deep}, dual DFXP with DFXP \cite{lin2021hybrid}, FXP with Posit \cite{gohil2021fixed}, BFP with LNS \cite{ni2020lbfp}, Posit with LNS \cite{murillo2021plam}, and RNS with LNS \cite{arnold2019implementing}. Combining two number systems allows for gaining from the benefits offered by both systems. The hybrid representations are found to be more efficient, a hardware and accuracy point of view, than using each representation separately. More combinations of these representations can be investigated in the future. For example, applying the same concept of BFP (i.e., each block shares the same exponent) to Posit number system is expected to relieve the hardware complication compared to the vanilla Posit number system.

\subsection{Utilization of DNN Characteristics}

DNN has special characteristics that should be considered when searching for more efficient representations dedicated to DNNs. For example, the ability of the neural networks to tolerate the noise is exploited in \cite{ansari2020improved} to design an efficient LNS multiplier by reducing the average rather than the absolute error introduced by the multiplier. This results in enhancing the accuracy of DNN instead of ruining it as would be anticipated when using approximate multipliers. Another example of utilizing the noise tolerance of DNNs is using stochastic rounding (i.e., rounding the number up or down at random) when the real number is mapped to a specific representation. This kind of rounding allowed for training DNNs with lower precision when it is integrated with FXP \cite{lin2016overcoming, gupta2015deep, chen2017fxpnet}, BFP \cite{zhang2021fast}, Posit \cite{cococcioni2019fast}, or DFXP \cite{gysel2016hardware}. Similarly, the ability to cluster DNN data into groups with narrower dynamic ranges gave birth to BFP and DFP representations. Moreover, realizing that DNN data are normally distributed shed light on the effectiveness of using the Posit number system, which has tapered accuracy. For future work on DNN number systems, these and other DNN characteristics should be paid attention to achieve more efficient representations. 

\section{Summary and Conclusions}\label{conclusions}
Deep neural networks have become an enabling component for a myriad of artificial intelligence applications. Being successful in providing great performance and even exceeding human accuracy, they have attracted the attention of academia and industry. The great performance of DNNs comes at the expense of high computational complexity and intensive memory requirements. Thus, increased attention is paid to redesigning DNN algorithms and hardware, in an effort to enhance their performance and/or enable their deployment on edge devices. A research direction that has a great impact on the performance of DNNs is their number representation. A great body of research has been focused on finding more suitable number systems, than FLP and FXP, tailored for DNNs. 

The standard FLP representation has a massive dynamic range which makes it a good choice for computationally intensive algorithms that include a wide range of values and require high precision. At the same time, the complex and power-hungry FLP calculations make it less attractive for DNN architecture implementation. On the other hand, the FXP for DNN implementation offers great hardware efficiency at the expense of accuracy degradation. Between the two extreme representations (FLP and FXP), there are several number systems that are used for DNNs and offer different trade-offs between energy efficiency and acquired accuracy. The surveyed alternative number systems for DNNs are LNS, RNS, BFP, DFXP, and Posit number systems.

The main objective of using LNS is to simplify the implementation of the costly multiplication operation and have a multiplication-free DNN accelerators. This hardware simplification allows for significant savings in the area, power consumption, and cost, with some accuracy degradation \footnote{This is the common case. However, several works that adopted LNS showed no accuracy degradation. See Table \ref{lns_multipliers_comp} and Table \ref{lns_quant_comp}.} resulting from logarithmic approximation. This makes LNS a good choice when DNNs are deployed on source-constrained devices for accuracy-resilience applications.

The RNS can be an attractive number representation choice for DNN accelerators. RNS exhibits inherent parallelism at the residue-processing level. It utilizes parallel computations along separate residue channels, where operations in each of them are performed modulo a specific modulus, with no need for information to be shared between residue channels. The main challenge in designing an efficient RNS-based accelerator is to minimize or, possibly, eliminate the overhead introduced when implementing the non-linear DNN operations. Another key factor is the optimization of the moduli selection and the corresponding arithmetic circuits, to meet the accuracy requirements.

The BFP strikes a balance between FLP and FXP format. Consequently, different trade-offs can be obtained by specifying different BFP design choices represented by the block size, shared exponent selection, and bit-width choice. Most of the surveyed DNN architectures that depend on BFP achieved negligible accuracy degradation compared to FLP even with less than 8 bits, with varying levels of speed, power, and area efficiency. DFXP can be considered as a subset of BFP with less dynamic range and less hardware complication at the same time. While BFP is closer to FLP, the DFXP is more like the FXP (as their names indicate). This results in different trade-offs between DNNs metrics (accuracy, power consumption, speed up, etc.)

Finally, Posit representation can be considered as a variant of FLP. offering better accuracy and a wider dynamic range, Posit can represent DNN data more efficiently with the same number of bits as FLP. This allows for more reduction in the number of bits compared to FLP implementations with similar accuracy. However, Posit has complex hardware, due to the hardware needed to convert Posit numbers to another number system (basically FLP) to do the arithmetic operations in the other domain before returning back to Posit domain. The efforts made to enhance its hardware efficiency have been discussed in this survey.

For all aforementioned alternative number systems, their impact on the performance and hardware design of DNN has been reported in details. In addition, this article highlighted the challenges associated with the implementation of each number system and the different solutions proposed to address these challenges. 

\section*{Acknowledgments}
This work was supported by the Khalifa University of Science and Technology under Award CIRA-2020-053.

\renewcommand{\baselinestretch}{1.4}
\bibliographystyle{IEEEtran}
\bibliography{IEEEabrv,references}

\end{document}